\documentclass{article}

\usepackage[final]{corl_2019} %

\usepackage{amssymb}
\usepackage{bbm}
\usepackage{amsmath}
\usepackage{pgfplots}
\usepackage[normalem]{ulem}
\usepackage{dsfont}
\usepackage{verbatim}
\usepackage{xspace}
\usepackage{blindtext}
\usepackage{wrapfig}
\usepackage{float}

\usepackage{multirow}

\usepackage{titlesec}
\titlespacing*{\section} {0pt}{0.9ex plus 0.1ex minus .1ex}{0.6ex plus .1ex}
\titlespacing{\paragraph} {0pt}{0ex plus 0ex minus .1ex}{0.5em}
\usepackage{caption}
\usepackage{arydshln}

\newcommand{\ddline}[1]{\cdashline{#1}[0.5pt/1pt]}

\usepackage{algorithmicx} 
\usepackage[noend]{algpseudocode}
\usepackage{algorithm}
\usepackage[T1]{fontenc}

\algnewcommand\algorithmicdefinitions{\textbf{Definitions:}}
\algnewcommand\Definitions{\item[\algorithmicdefinitions]}
\renewcommand{\algorithmiccomment}[1]{{\color{gray}\raisebox{1px}{\texttt{\guillemotright}} #1}}
\newcommand{\Commentx}[1]{{\color{gray}\phantom{\raisebox{1px}{\texttt{\guillemotright}}} #1}}
\algnewcommand{\LineComment}[1]{\Statex \hskip\ALG@thistlm \algorithmiccomment{#1}}
\algrenewcommand\alglinenumber[1]{\footnotesize #1:}
\algrenewcommand\algorithmicindent{1.0em}%
\makeatletter
\newcommand{\StatexIndent}[1][3]{%
  \setlength\@tempdima{\algorithmicindent}%
  \Statex\hskip\dimexpr#1\@tempdima\relax}

\makeatletter
\g@addto@macro\small{%
  \setlength\abovedisplayskip{-13pt}
  \setlength\abovedisplayshortskip{-13pt}
  \setlength\belowdisplayshortskip{-9pt}
  \setlength\belowdisplayskip{-9pt}
}
\makeatother

\newcommand{\suploss}{\mathcal{L}_{\rm SL}}

\newcommand{\domainsim}{\texttt{S}}
\newcommand{\domainreal}{\texttt{R}}

\newcommand{\posoob}{p^{\texttt{oob}}}
\newcommand{\dataset}{\mathcal{D}}

\newcommand{\allpositions}{\mathcal{P}}
\newcommand{\observedpositions}{\mathcal{P}^{\texttt{obs}}}

\newcommand{\reward}{r}

\newcommand{\stopreward}{\reward_{s}}
\newcommand{\visitreward}{\reward_{v}}

\newcommand{\actionreward}{\reward_{a}}
\newcommand{\explorereward}{\reward_{e}}

\newcommand{\stopweight}{\lambda_{s}}
\newcommand{\visitweight}{\lambda_{v}}
\newcommand{\stepweight}{\lambda_{step}}
\newcommand{\actionweight}{\lambda_{a}}
\newcommand{\exploreweight}{\lambda_{e}}

\newcommand{\potential}{\phi}
\newcommand{\visitpotential}{\potential_{v}}
\newcommand{\explorepotential}{\potential_{e}}

\newcommand{\paramsA}{\theta}
\newcommand{\paramsAsim}{\theta_{\domainsim}}
\newcommand{\paramsAreal}{\theta}
\newcommand{\paramsB}{\phi}
\newcommand{\stageA}{f}
\newcommand{\stageAExpert}{f^*}
\newcommand{\stageAsim}{f_{\domainsim}}
\newcommand{\stageAreal}{f}
\newcommand{\stageB}{g}
\newcommand{\discriminator}{h}
\newcommand{\discrimparams}{\psi}
\newcommand{\valueparams}{\upsilon}
\newcommand{\valuefunc}{V}
\newcommand{\clamp}{T_{\psi}}

\newcommand{\eat}[1]{\ignorespaces}

\newcommand{\instrlen}{l}
\newcommand{\execlen}{T}
\newcommand{\idxtimestep}{t}

\newcommand{\trainsetsize}{N}
\newcommand{\testsetsize}{M}

\newcommand{\reals}{\mathds{R}}

\newcommand{\state}{s}

\newcommand{\configuration}{\rho}
\newcommand{\position}{p}
\newcommand{\posseq}{\Xi}
\newcommand{\execposseq}{\hat{\Xi}}

\newcommand{\execution}{\Xi}
\newcommand{\goalpos}{\position_g}

\newcommand{\trajvisit}{d^{\position}}
\newcommand{\stopvisit}{d^{g}}

\newcommand{\emd}{\textsc{EMD}}

\newcommand{\context}{c}
\newcommand{\gencontexts}{\mathcal{C}}

\newcommand{\act}[1]{{\tt #1}}
\newcommand{\action}{a}

\newcommand{\stopaction}{\act{STOP}}

\newcommand{\nlstring}[1]{{\em #1}}

\newcommand{\rail}{\textsc{SuReAL}\xspace}
\newcommand{\fullrail}{Supervised and Reinforcement Asynchronous Learning\xspace}

\newcommand{\instruction}{u}
\newcommand{\instructionemb}{\mathbf{u}}

\newcommand{\params}{\theta}

\newcommand{\velfwd}{v}
\newcommand{\velang}{\omega}
\newcommand{\stopprob}{p^{\stopaction}}
\newcommand{\image}{I}
\newcommand{\pose}{P}
\newcommand{\worldframe}{W}
\newcommand{\camframe}{C}

\newcommand{\fmcam}{\mathbf{F}^\camframe}

\newcommand{\fmworld}{\mathbf{F}^\worldframe}
\newcommand{\maskworld}{\mathbf{M}^\worldframe}
\newcommand{\boundaryworld}{\mathbf{B}^\worldframe}
\newcommand{\smworld}{\mathbf{S}^\worldframe}

\newcommand{\rmworld}{\mathbf{R}^\worldframe}

\newcommand{\weights}{\mathbf{W}}
\newcommand{\bias}{\mathbf{b}}

\newcommand{\resnet}{\textsc{CNN} }

\newcommand{\lingunet}{\textsc{LingUNet}}
\newcommand{\unet}{\textsc{UNet}}

\newcommand{\oraclemodel}{\textsc{Oracle}}
\newcommand{\stopmodel}{\textsc{Stop}}
\newcommand{\avgmodel}{\textsc{Average}}

\newcommand{\featmap}{\mathbf{F}}
\newcommand{\featmaptxt}{\mathbf{G}}
\newcommand{\featmapdeconv}{\mathbf{H}}
\newcommand{\conv}{\textsc{CNN}}

\newcommand{\upscale}{\textsc{Upscale}}

\newcommand{\lingunetvecout}{\textbf{h}}
\newcommand{\avgpool}{\textsc{AvgPool}}
\newcommand{\kernel}{\mathbf{K}}

\title{Learning to Map Natural Language Instructions to Physical Quadcopter Control using Simulated Flight}

 \newcommand{\authorgap}{\hspace{1em}}
 \author{
   Valts Blukis$^{1}$ \authorgap Yannick Terme$^{2}$ \authorgap Eyvind Niklasson$^{3}$ \authorgap Ross A. Knepper$^{4}$ \authorgap Yoav Artzi$^{5}$ \vspace{0.5em}\\
   $^{1,4,5}$Department of Computer Science, Cornell University, Ithaca, New York, USA\\
 $^{1,2,3,5}$Cornell Tech, Cornell University, New York, New York, USA\\
 	\texttt{\{$^1$valts, $^4$rak, $^5$yoav\}@cs.cornell.edu} \authorgap \texttt{$^2$yannickterme@gmail.com} 	\\
 	\texttt{$^3$een7@cornell.edu}
 }

\begin{document}

\maketitle
\vspace{-1.6em}

\begin{abstract}
We propose a joint simulation and real-world learning framework for mapping navigation instructions and raw first-person observations to continuous control. Our model estimates the need for environment exploration, predicts the likelihood of visiting environment positions during execution, and controls the agent to both explore and visit high-likelihood positions. We introduce Supervised Reinforcement Asynchronous Learning (SuReAL). Learning uses both simulation and real environments without requiring autonomous flight in the physical environment during training, and combines supervised learning for predicting positions to visit and reinforcement learning for continuous control. We evaluate our approach on a natural language instruction-following task with a physical quadcopter, and demonstrate effective execution and exploration behavior. 

\end{abstract}


\section{Introduction}
\label{sec:introduction}

Controlling robotic agents to execute natural language instructions requires addressing perception, language, planning, and control challenges. 
The majority of methods addressing this problem follow such a decomposition, where separate components are developed independently and are then combined together~\cite[e.g.,][]{tellex11grounding, matuszek2012joint, duvallet2013imitation, walter2013learning, hemachandra2015learning, gopalan2018sequence}. 
This requires a hard-to-scale engineering intensive process of designing and working with intermediate representations, including a formal language to represent natural language meaning.
Recent work instead learns intermediate representations, and uses a single model to address all reasoning challenges~\cite[e.g.,][]{lenz2015deep, levine2016learning, sadeghi2016cad2rl, quillen2018deep}.
So far, this line of work has mostly focused on pre-specified low-level tasks. In contrast, executing natural language instructions requires understanding sentence structure, grounding words and phrases to observations, reasoning about previously unseen tasks, and handling ambiguity and uncertainty.

\begin{figure*}[t]
\centering
\includegraphics[scale=0.305]{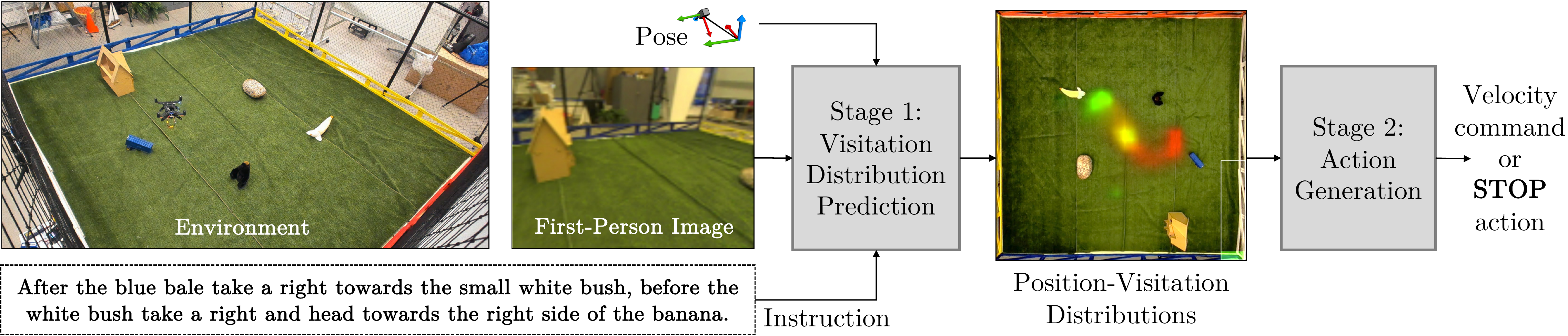}
\caption{
An illustration of our task and model. Correct execution of the instruction requires recognizing objects (e.g., \nlstring{blue bale}), inferring a path (e.g., to turn to the right \nlstring{after the blue bale}), and  generating the commands to steer the quadcopter and stop at the goal location.
The model input at time $\idxtimestep$ is the instruction $\instruction$, a first-person RGB observation $\image_\idxtimestep$, and a pose estimate $\pose_\idxtimestep$. 
The model has two stages: predicting the probability of visiting positions during execution and generating actions. 
}
\label{fig:task}
\vspace{20pt}
\centering
\includegraphics[scale=0.297]{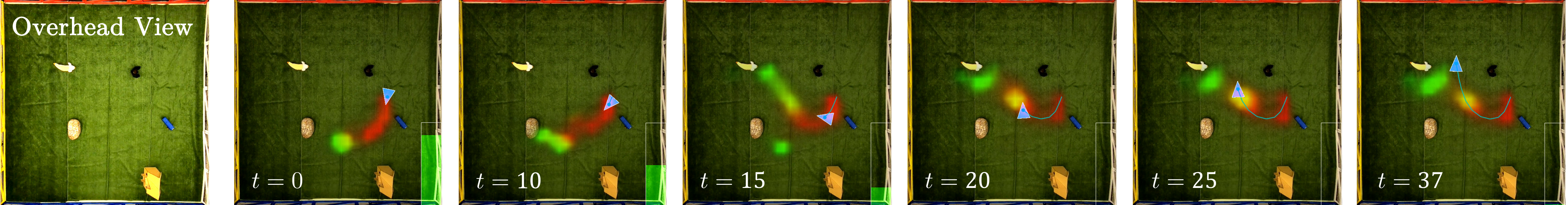}
\caption{
Predicted visitation distributions as the instruction execution progresses (left-to-right), trajectory in red and goal in green, for the example from Figure~\ref{fig:task}. The green bar shows the agent's belief  the goal has not been observed yet. 
A video of the execution and intermediate representations is available at \href{https://youtu.be/PLdsNPE4Gz4}{https://youtu.be/PLdsNPE4Gz4}.
}
\label{fig:exec}
\end{figure*}

In this paper, we address the problem of mapping natural language navigation instructions to continuous control of a quadcopter drone using representation learning. 
We present a neural network model to jointly reason about observations, natural language, and robot control, with explicit modeling of the agent's plan and exploration of the environment. 
For learning, we introduce \fullrail ($\rail$), a method for joint training in simulated and physical environments. 
Figure~\ref{fig:task} illustrates our task and model. 

We design our model to reason about 
partial observability and incomplete knowledge of the environment in instruction following.
We explicitly model observed and unobserved areas, and the agent's belief that the goal location implied by the instruction has been observed. 
During learning, we use an intrinsic reward to encourage behaviors that increase this belief, and penalize for indicating task completion while still believing the goal is unobserved. 

$\rail$ addresses two key learning challenges. 
First, flying in a physical environment at the scale needed  for our complex learning task is both time-consuming and costly. 
We mitigate this problem using a simulated environment. 
However, in contrast to the common approach of domain transfer from simulated to physical environments~\cite{rusu2017sim, bousmalis2017using}, we simultaneously train in both, while not requiring autonomous flight in the physical environment during training. 
Second, as each example requires a human instruction, it is prohibitively expensive to collect language data at the scale required for representation learning~\cite{hermann2017grounded, chaplot2017gated}. This is unlike tasks where data can be collected without human interaction. 
We combine supervised and reinforcement learning (RL); the first to best use the limited language data, and the second to effectively leverage experience.   
We evaluate our approach with a navigation task, where a quadcopter drone flies between landmarks following natural language instructions. 
We modify an existing natural language dataset~\cite{misra2018mapping} to create a new benchmark with long instructions, complex trajectories, and observability challenges. 
We evaluate using both automated metrics and human judgements of semantic correctness. 
To the best of our knowledge, this is the first demonstration of a physical quadcopter system that follows natural language instructions by mapping raw first-person images and pose estimates to continuous control. 
Our code, data, and demo videos are available at \href{https://github.com/lil-lab/drif}{https://github.com/lil-lab/drif}.

\section{Technical Overview}
\label{sec:tech_overview}

\paragraph{Task}
Our goal is to map natural language navigation instructions to continuous control of a quadcopter drone. 
The agent behavior is determined by a velocity controller setpoint $\configuration = ( \velfwd, \velang )$, where  $\velfwd\in\reals$ is a forward velocity  and  $\velang\in\reals$ is a yaw rate.
The model generates actions at fixed intervals.  
An 
action is either the task completion action $\stopaction$ or a setpoint update $( \velfwd, \velang ) \in \reals^2$. 
Given a setpoint update $\action_{\idxtimestep} = (\velfwd_{\idxtimestep}, \velang_{\idxtimestep})$ at time $\idxtimestep$, we fix the controller setpoint $\configuration = (\velfwd_{\idxtimestep}, \velang_{\idxtimestep})$ that is maintained between actions.
Given a start state $\state_1$ and an instruction $\instruction$, an execution $\execution$ of length $\execlen$ is a sequence  $\langle (\state_1, \action_1), \dots, (\state_{\execlen}, \action_{\execlen}) \rangle$, where $\state_\idxtimestep$ is the state at time $\idxtimestep$, $\action_{\idxtimestep<\execlen} \in \reals^2$ are setpoint updates, and $\action_{\execlen} = \stopaction$. 
The state includes the quadcopter pose, internal state, and all landmark locations.

\paragraph{Model}

We assume access to raw first-person monocular observations and pose estimates. 
The agent does not have access to the world state. 
At time $\idxtimestep$, the agent observes the \emph{agent context} $\context_{\idxtimestep} = (\instruction, \image_1,\cdots, \image_{\idxtimestep}, \pose_1, \cdots \pose_{\idxtimestep})$,  where $\instruction$ is the instruction and $\image_i$ and $\pose_i$ are monocular first-person RGB images and 6-DOF agent poses observed at time $i$. 
We base our model on the Position Visitation Network~\cite[PVN;][]{blukis2018mapping} architecture, and introduce mechanisms for reasoning about observability and exploration and learning across simulated and real environments. 
The model operates in two stages: casting 
planning as predicting distributions over world positions indicating the probability of visiting a position during execution, and generating actions to visit high probability positions.

\paragraph{Learning}

We train jointly in simulated and physical environments. 
We assume access to a simulator and demonstration sets in both environments, $\dataset^{\domainreal}$ in the physical environment and $\dataset^{\domainsim}$ in the simulation. 
We do not interact with the physical environment during training.
Each dataset includes $\trainsetsize^{D}$ examples $\{ (\instruction^{(i)},  \execution^{(i)})\}_{i = 1}^{\trainsetsize^{D}}$, where $D\in\{\domainreal,\domainsim\}$, $\instruction^{(i)}$ is an instruction, and $\execution^{(i)}$ is a demonstration execution.
We do not require the datasets to be aligned or provide demonstrations for the same set of instructions. 
We propose $\rail$, a learning approach  that concurrently trains the two model stages in two separate processes. The planning stage is trained with supervised learning, while the action generation stage is trained with RL. 
The two processes exchange data and parameters. The trajectories collected during RL are added to the dataset used for supervised learning, and the planning stage parameters are periodically transferred to the RL process training the action generation stage. 
This allows the action generator to learn to execute the plans predicted by the planning stage, which itself is trained using on-policy observations collected from the action generator.

\paragraph{Evaluation}
We evaluate  on a test set of $\testsetsize$ examples $\{ (\instruction^{(i)}, \state_1^{(i)},  \posseq^{(i)}) \}_{i = 1}^{\testsetsize}$, where $\instruction^{(i)}$ is an instruction, $\state_1^{(i)}$ is a start state, and $\posseq^{(i)}$ is a human demonstration. 
We use human evaluation to verify the generated trajectories are semantically  correct with regard to the instruction. 
We also use automated metrics. 
We consider the task successful if the agent stops  within a predefined Euclidean distance of the final position in $\posseq^{(i)}$. 
We evaluate  the quality of generating the trajectory following the instruction using earth mover's distance between $\posseq^{(i)}$ and executed trajectories.

\section{Related Work}
\label{sec:related}

Natural language instruction following has been extensively studied using hand-engineered symbolic intermediate representations of world state or instruction semantics with physical robots~\cite{tellex11grounding, matuszek2012joint, duvallet2013imitation, walter2013learning, misra2014context, hemachandra2015learning, thomason2015learning, gopalan2018sequence, williams2018learning} and simulated agents~\cite{macmahon2006walk, branavan2010reading, matuszek2012learning, artzi2013weakly,Artzi:14, suhr2018situated}.
In contrast, we study trading off the symbolic representation design with representation learning from demonstrations.

Representation learning has been studied for executing specific tasks such as grasping~\cite{lenz2015deep, levine2016learning, quillen2018deep}, dexterous manipulation~\cite{levine2016end, nair2017combining}, or continuous flight~\cite{sadeghi2016cad2rl}. 
Our aim is to execute navigation tasks specified in natural language, including new tasks at test time. 
This problem was addressed with representation learning in discrete simulated environments~\cite{misra2017mapping, shah2018follownet, anderson2017vision, misra2018mapping, fried2018speaker, jain2019stay}, and more recently with continuous simulations~\cite{blukis2018mapping}. 
However, these methods were not demonstrated on physical robots. 
A host of problems combine to make this challenging, including grounding natural language to constantly changing observations, robustly bridging the gap between relatively high-level instructions to continuous control, and learning with limited language data and the high costs of robot usage. 
Our model is based on the Position Visitation Network~\cite{blukis2018mapping} architecture that incorporates geometric computation to represent language and observations in learned spatial maps. 
This approach is related to neural network models that construct maps~\cite{gupta2017cognitive, blukis2018following, khan2018memory, anderson2019chasing} or perform planning~\cite{srinivas2018universal}.

Our approach is aimed at a partial observability scenario and does not assume access to the complete system state. 
Understanding the instruction often requires identifying mentioned entities that are not initially visible. This requires exploration during task execution. 
\citet{nyga2018grounding} studied modeling incomplete information in instructions with a modular approach. In contrast, we jointly learn to infer the absence of necessary information and to remedy it via exploration.

\section{Model}
\label{sec:model}

\begin{figure*}[t]
\centering
\includegraphics[scale=0.334]{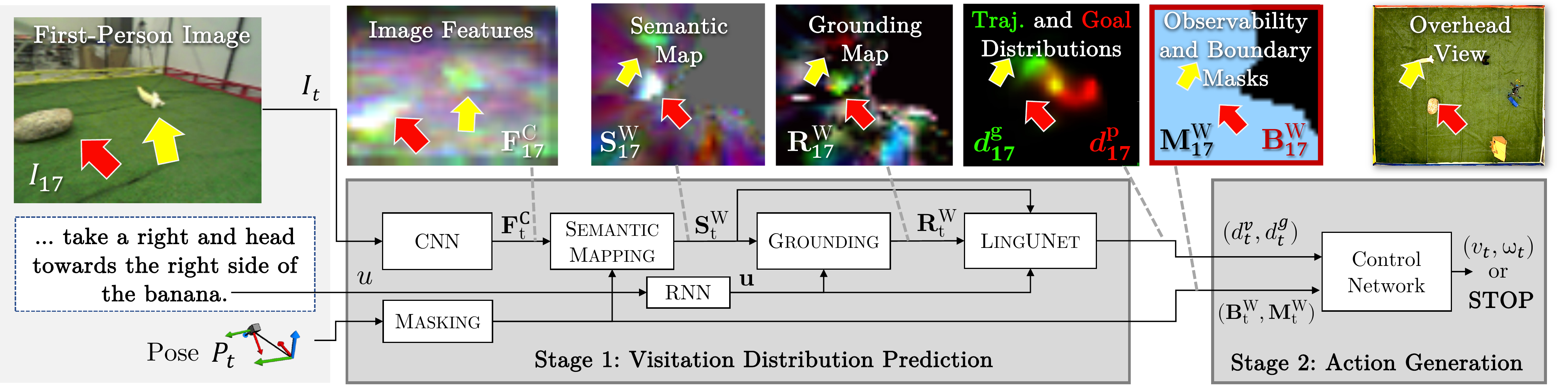}
\caption{Model architecture illustration. The first stage generates a semantic map $\smworld$, a grounding map $\rmworld$, observability masks $\maskworld_\idxtimestep$ and $\boundaryworld_\idxtimestep$, and visitation distributions $\trajvisit_t$ and $\stopvisit_t$. The red and yellow arrows indicate the rock and banana locations. We show all intermediate representations at timestep 17 out of 37, after most of the environment has been observed. Figure~\ref{fig:exec} shows the visitation distributions for other timesteps, and Figure~\ref{fig:exec_repr} in the appendix shows all timesteps.
An animated version of this figure is available at \href{https://youtu.be/UuZtSl6ckTk}{https://youtu.be/UuZtSl6ckTk}.}
\label{fig:model}
\end{figure*}

We model the policy $\pi$ with a neural network. At time $\idxtimestep$, given the  agent context $\context_\idxtimestep$, the policy outputs a stopping probability $\stopprob_\idxtimestep$, a forward velocity $\velfwd_{\idxtimestep}$, and an angular velocity $\velang_{\idxtimestep}$. 
We decompose the architecture to two stages $\pi(\context_{\idxtimestep}) = \stageB(\stageAreal(\context_{\idxtimestep}))$, where $\stageAreal$ predicts the probability of visiting positions  in the environment and $\stageB$ generates the actions to visit high probability positions. 
The position visitation probabilities are continuously updated during execution to incorporate the most recent observations, and past actions directly affect the information available for future decisions.
Our model is based on the PVN architecture~\cite{blukis2018mapping}. 
We introduce several improvements, including explicit modeling of observability in both stages. 
Appendix~\ref{app:sec:model} contains further model implementation details, including a detailed list of our improvements.
Figure~\ref{fig:model} illustrates our model for an example input.

\paragraph{Stage 1: Visitation Distribution Prediction}

At time $t$, the first stage $\stageA(\cdot)$ generates two probability distributions: a trajectory-visitation distribution $\trajvisit_t$ and a goal-visitation distribution $\stopvisit_t$. 
Both distributions assign probabilities to positions $\observedpositions \cup \{\posoob\}$, where $\observedpositions$ is the set of  positions observed up to time $t$ and $\posoob$  represents all yet-unobserved positions. 
The set $\observedpositions$ is a discretized approximation of the continuous environment. This approximation enables efficient computation of the visitation distributions~\cite{blukis2018mapping}. 
The trajectory-visitation distribution $\trajvisit$ assigns high probability to positions  the agent should go through during execution, and the goal-visitation distribution $\stopvisit$ puts high probability on positions where the agent should $\stopaction$ to complete its execution. 
We add the special position $\posoob$ to $\textsc{PVN}$ to capture probability mass that should otherwise be assigned to positions not yet observed, for example when the goal position has not been observed yet.

The first stage combines a learned neural network and differentiable deterministic computations. 
The input instruction $\instruction$ is mapped to a vector $\instructionemb$ using a recurrent neural network (RNN). The input image $\image_\idxtimestep$  is mapped to a feature map $\fmcam_\idxtimestep$ that captures spatial and semantic information using a convolutional neural network (CNN). 
The feature map $\fmcam_\idxtimestep$ is processed using a deterministic semantic mapping process~\cite{blukis2018following} using a pinhole camera model and the agent pose $\pose_\idxtimestep$ to project $\fmcam_\idxtimestep$ onto the environment ground at zero elevation. 
The projected features are deterministically accumulated from previous timesteps to create a semantic map $\smworld_\idxtimestep$. 
$\smworld_\idxtimestep$ represents each position with a learned feature vector aggregated from all past observations of that position.
For example, in Figure~\ref{fig:model}, the \nlstring{banana} and the \nlstring{white bush} can be identified in the raw image features $\fmcam_\idxtimestep$ and the projected semantic map $\smworld_\idxtimestep$, where their representations are identical. 
We generate a language-conditioned grounding map $\rmworld_\idxtimestep$ by creating convolutional filters using the text representation $\instructionemb$ and filtering $\smworld_\idxtimestep$. 
The two maps aim to provide different representation: $\smworld_\idxtimestep$ aims for a language-agnostic environment representation and $\rmworld_\idxtimestep$ is intended to focus on the objects mentioned in the instruction.  
We use auxiliary objectives (Appendix~\ref{sec:app:aux}) to optimize each map to contain the intended information. 
We predict the two distributions using $\lingunet$~\cite{misra2018mapping}, a language-conditioned variant of the $\unet$ image reconstruction architecture~\cite{ronneberger2015u}, 
which takes as input the learned maps, $\smworld_\idxtimestep$ and $\rmworld_\idxtimestep$, and the instruction representation $\instructionemb$. 
Appendix~\ref{app:sec:model:stage1} provides a detailed description of this architecture. 

We add two outputs to the original PVN design: an observability mask $\maskworld_\idxtimestep$ and a boundary mask $\boundaryworld_\idxtimestep$. 
Both are computed deterministically given the agent pose estimate and the camera parameters, and are intended to aid exploration of the environment during instruction execution.
$\maskworld_\idxtimestep$ assigns $1$ to each position $\position \in \allpositions$ in the environment if $\position$ has been observed by the agent's first-person camera by time $\idxtimestep$, or $0$ otherwise. $\boundaryworld_\idxtimestep$ assigns $1$ to environment boundaries and $0$ to other positions. Together, the masks provide information about what parts of the environment remain to be explored.

\paragraph{Stage 2: Action Generation}

The second stage $\stageB(\cdot)$ is a \emph{control network} that receives four inputs: a trajectory-visitation distribution $\trajvisit_\idxtimestep$, a goal-visitation visitation distribution $\stopvisit_\idxtimestep$, an observability mask $\maskworld_\idxtimestep$, and a boundary mask $\boundaryworld_\idxtimestep$. 
The four inputs are rotated to the current egocentric agent reference frame, and used to generate the output velocities using a learned neural network. 
Appendix~\ref{sec:app:control} describes the network architecture. 
The velocities are generated to visit high probability positions according to $\trajvisit$, and the $\stopaction$ probability is predicted to stop in a likely position according to $\stopvisit$. 
In the figure, $\trajvisit_t$ shows the curved flight path, and $\stopvisit_t$ identifies the goal \nlstring{right of the banana}. 
Our addition of the two masks and the special position $\posoob$ enables generating actions to explore the environment to reduce $\trajvisit(\posoob)$ and $\stopvisit(\posoob)$. Figure~\ref{fig:exec} visualizes $\stopvisit(\posoob)$ with a green bar, showing how it starts high, and decreases once the goal is observed at step $t=15$.

\section{Learning}
\label{sec:learning}

We learn two sets of parameters: $\paramsA$ for the first stage $\stageA(\cdot)$ and $\paramsB$ for the second stage $\stageB(\cdot)$. 
We use a simulation for all autonomous flight during learning, and jointly train for both the simulation and the physical environment. 
This includes training a simulation-specific first stage $\stageAsim(\cdot)$ with additional parameters $\paramsAsim$. The second stage model $\stageB(\cdot)$ is used in both environments. 
We assume access to sets of training examples $\dataset^{\domainsim} = \{ (\instruction^{(i)},  \execution^{(i)})\}_{i = 1}^{\trainsetsize^{\domainsim}}$ for the simulation and  $\dataset^{\domainreal} = \{ (\instruction^{(i)},  \execution^{(i)})\}_{i = 1}^{\trainsetsize^{\domainreal}}$ for the physical environment, where $\instruction^{(i)}$ are instructions and $\execution^{(i)}$ are demonstration executions. The training examples are not spatially or temporally aligned between domains.

\begin{figure*}[t]
\centering
\includegraphics[scale=0.285]{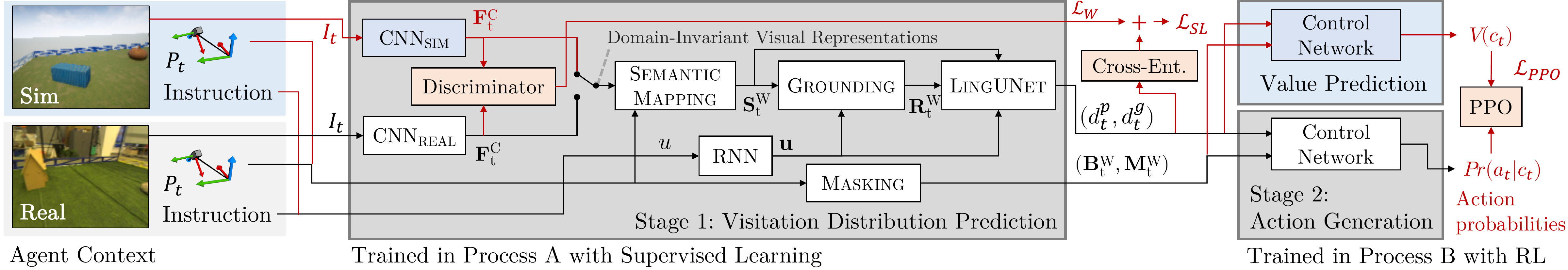}
\caption{Learning architecture. Stages 1 and 2 of our  model are concurrently trained in processes A and B. The blue and orange blocks, and red arrows, represent modules, computation, and loss functions during training only. The white blocks form the final learned policy. We learn from inputs from simulation and real world environments, by switching between the two $\resnet$ modules.}
\label{fig:learning_arch}
\end{figure*}

Our learning algorithm, \fullrail ($\rail$), uses two concurrent asynchronous processes. 
Each process only updates the parameters of one  stage. 
Process A uses supervised learning to estimate Stage 1 parameters for both environments: $\paramsAreal$ for the physical environment model $\stageAreal(\cdot)$ and $\paramsAsim$ for $\stageAsim(\cdot)$. 
We use both $\dataset^\domainreal$ and $\dataset^\domainsim$ to update the model parameters.   
We use RL in Process B to learn the parameters $\paramsB$ of the second stage $\stageB(\cdot)$ using an intrinsic reward function. 
We start learning using the provided demonstrations in $\dataset^\domainsim$ and periodically replace execution trajectories with RL rollouts from Process B, keeping a single execution per instruction at any time. We warm-start by running Process A for $K_{\rm iter}^{B}$ iterations before launching Process B to make sure that Process B always receives as input sensible visitation predictions instead of noise.
The model parameters are  periodically synchronized by copying the simulation parameters of Stage 1 from Process A to B. 
For learning, we use a learning architecture (Figure~\ref{fig:learning_arch}) that extends our model to process simulation observations and adds a discriminator that encourages  learning representations that are invariant to the type of visual input.

\paragraph{Process A: Supervised Learning for Visitation Prediction}

We train $\stageAreal(\cdot)$ and $\stageAsim(\cdot)$ to: (a) minimize the KL-divergence between the predicted visitation distributions and reference distributions generated from the demonstrations, and (b) learn domain invariant visual representations that allow sharing of instruction grounding and execution between the two environments. 
We use demonstration executions in the real environment $\dataset^\domainreal$ and in the simulated environment $\dataset^\domainsim$. 
The loss for executions from the physical environment $\execution^\domainreal$ and the simulation $\execution^\domainsim$ is: 

\begin{small}
\begin{equation}
 \suploss(\execution^\domainreal, \execution^\domainsim) = \frac{1}{|\execution^\domainreal|}\sum_{\context \in \gencontexts(\execution^\domainreal)}D_{\rm KL}(\stageAreal(\context) \| \stageAExpert(\context))  +
    \frac{1}{|\execution^\domainsim|}\sum_{\context \in \gencontexts(\execution^\domainsim)}D_{\rm KL}(\stageAsim(\context) \| \stageAExpert(\context)) +
    \mathcal{L}_{W}(\execution^\domainreal, \execution^\domainsim)\;\;,
\label{eq:wass}
\end{equation}
\end{small}

where $\gencontexts(\execution)$ is the sequence of contexts observed by the agent during an execution $\execution$ and $\stageAExpert(\context)$ creates the gold-standard visitation distribution examples (i.e., Stage 1 outputs) for a context $\context$ from the training data. 
The term $\mathcal{L}_{W}(\execution^\domainreal, \execution^\domainsim)$ aims to make the feature representation $\fmcam$ indistinguishable between real and simulated images. This allows the rest of the model to use either simulated or real observations interchangeably. 
$\mathcal{L}_{W}(\execution^\domainreal, \execution^\domainsim)$ is the approximated empirical Wasserstein distance between the visual feature distributions extracted from simulated and real agent contexts:

\begin{small}
\begin{equation*}
\mathcal{L}_{W}(\execution^\domainreal, \execution^\domainsim) = \frac{1}{|\execution^\domainsim|}\sum_{\context_\idxtimestep \in \execution^\domainsim}\discriminator(\resnet^\domainsim(\image_\idxtimestep)) - \frac{1}{|\execution^\domainreal|}\sum_{\context_\idxtimestep \in \execution^\domainreal}\discriminator(\resnet(\image_\idxtimestep))\;\;,
\end{equation*}
\end{small}

where $\discriminator$ is a Lipschitz continuous neural network discriminator with parameters $\discrimparams$ that we train to output high values for simulated features and low values for real features~\cite{shen2017wasserstein,arjovsky2017wasserstein}. 
$\image_\idxtimestep$ is the $\idxtimestep$-th image in the agent context $\context_\idxtimestep$. 
The discriminator architecture is described in Appendix~\ref{app:discriminator}.

Algorithm~\ref{algo:supervised} shows the supervised optimization procedure. 
We alternate between updating the discriminator parameters $\discrimparams$, and the first stage model parameters $\paramsAreal$ and $\paramsAsim$.
At every iteration, we perform $K^{\rm SL}_{\rm discr}$ gradient updates of $\discrimparams$ to maximize the Wasserstein loss $\mathcal{L}_W$ (lines \ref{ln:sl:discrloop}--\ref{ln:sl:discrupdate}), and then perform a single gradient update to $\paramsAreal$ and $\paramsAsim$ to minimize supervised learning loss $\mathcal{L}_{\rm SL}$ (lines \ref{ln:sl:modelupdatetraj}--\ref{ln:sl:modelupdate}).
We send the simulation-specific parameters $\paramsAsim$ to the RL process every $K^{\rm SL}_{\rm iter}$ iterations (line~\ref{ln:sl:updateA}).

\begin{figure}[t]
\centering
\begin{minipage}[t]{0.465\textwidth}
\begin{algorithm}[H]
\caption{Process A: Supervised Learning}
\begin{algorithmic}[1]
\footnotesize
\Require First stage models $\stageAreal$ and $\stageAsim$  with parameters $\paramsAreal$ and $\paramsAsim$, 
discriminator $\discriminator$ with parameters $\discrimparams$, 
datasets of simulated and physical environment trajectories ${\dataset^{\domainsim}}$ and $\dataset^{\domainreal}$. 
\Definitions $\dataset^\domainsim$ and $\stageAsim^B$ are shared with Process B.
\While{Process B has not finished}
    \For{$i = 1, \dots , K^{\rm SL}_{\rm iter}$}
        \For{$j = 1, \dots, K^{\rm SL}_{\rm discr}$}\label{ln:sl:discrloop}
            \State \Comment{Sample trajectories}
            \State $\execution^{\domainreal}\sim \dataset^{\domainreal}$ and $\execution^{\domainsim} \sim\dataset^{\domainsim}$
            \State \Comment{Update discriminator to maximize}
            \StatexIndent[3] \Commentx{Wasserstein distance}
            \State $\discrimparams \leftarrow \textsc{ADAM}(\nabla_{\discrimparams}-\mathcal{L}_{W}(\execution^{\domainreal},\execution^{\domainsim}))$ \label{ln:sl:discrupdate}
        \EndFor
        \State $\execution^{\domainreal}\sim \dataset^{\domainreal}$ and $\execution^{\domainsim} \sim\dataset^{\domainsim}$ \label{ln:sl:modelupdatetraj}
        \State \Comment{Update first stage parameters}
        \State $(\paramsAsim, \paramsAreal) \leftarrow \textsc{ADAM}(\nabla_{\paramsAsim,\paramsAreal}\suploss(\execution^\domainreal, \execution^\domainsim))$ \label{ln:sl:modelupdate}

    \EndFor
    \State \Comment{Send $\stageAsim$ to Process B if it is running}
    \State $\stageAsim^B \gets \stageAsim$ \label{ln:sl:updateA}
    \If{$i = K_{\rm iter}^{B}$}
        \State Launch Process B (Algorithm~\ref{algo:rl}) \label{ln:sl:launchB}
        \vspace{-2pt}
    \EndIf
\EndWhile\\ 
\Return $\stageAreal$
\vspace{-3pt}
\end{algorithmic}
\label{algo:supervised}
\end{algorithm}
\end{minipage}~
\begin{minipage}[t]{0.53\textwidth}
\begin{algorithm}[H]
\caption{Process B: Reinforcement Learning}
\begin{algorithmic}[1]
\footnotesize
\Require Simulation dataset ${\dataset^{\domainsim}}$, second-stage model $\stageB$ with parameters $\paramsB$, value function $\valuefunc$ with parameters $\valueparams$, first-stage simulation model $\stageAsim$. 
\Definitions $\textsc{Merge}(\dataset, E)$ is a set of sentence-execution pairs including all instructions from $\dataset$, where each instruction is paired with an execution from $E$, or $\dataset$ if not in $E$. $\dataset^\domainsim$ and $\stageAsim^B$ are shared with Process A.

\For{$e = 1,  \dots, K^{\rm RL}_{\rm epoch}$}
    \State \Comment{Get the most recent update from Process A}
    \State $\stageAsim \gets \stageAsim^B $
    \For{$i = 1,  \dots , K^{\rm RL}_{\rm iter}$}
        \State \Comment{Sample simulator executions of $N$ instructions}
        \State $\execposseq^{(1)}, ..., \execposseq^{(N)} \sim \stageB(\stageAsim(\cdot))$ \label{ln:rl:exec}
        \For{$j = 1, \dots, K^{\rm RL}_{\rm steps}$} \label{ln:rl:ppobegin}
            \State \Comment{Sample state-action-return tuples and update} 
            \State $X \sim \execposseq_{1}, ..., \execposseq_{N}$
            \State $\paramsB, \valueparams \leftarrow \textsc{ADAM}(\nabla_{\paramsB,\valueparams}\mathcal{L}_{PPO}(X,  \valuefunc))$
        \EndFor \label{ln:rl:ppoend}
        \State \Comment{Update executions to share with Process A}
        \State $\dataset^{\domainsim} \leftarrow \textsc{Merge}(\dataset^{\domainsim}, \{\execposseq_{1}, \dots , \execposseq_{N}\})$ \label{ln:rl:mergeexec}
    \EndFor
\EndFor\\
\Return $\stageB$
\end{algorithmic}
\label{algo:rl}
\end{algorithm}
\end{minipage}
\vspace{-12pt}
\end{figure}

\paragraph{Process B: Reinforcement Learning for Action Generation}

We train the action generator $\stageB(\cdot)$ using RL with an intrinsic reward.
We use Proximal Policy Optimization~\cite[PPO;][]{schulman2017proximal} to maximize the expected return.
The learner has no access to an external task reward, but instead computes a reward $\reward(\cdot)$ from how well the agent follows the visitation distributions generated by the first stage: 

\begin{small}
\begin{equation}
\reward(\context_\idxtimestep, \action_\idxtimestep) =
    \visitweight \visitreward(\context_\idxtimestep, \action_\idxtimestep) + \stopweight \stopreward(\context_\idxtimestep, \action_\idxtimestep) +  \exploreweight \explorereward(\context_\idxtimestep, \action_\idxtimestep) - 
    \actionweight \actionreward(\action_\idxtimestep) - \stepweight \nonumber\;\;,
\end{equation}
\end{small}

where $\context_\idxtimestep$ is the agent context at time $\idxtimestep$ and $\action_\idxtimestep$ is the action. 
The reward $\reward(\cdot)$ is a weighted combination of five terms. 
The visitation reward $\visitreward(\cdot)$ is the per-timestep reduction in earth mover's distance between the predicted distribution $\trajvisit_\idxtimestep$ and an empirical distribution that assigns equal probability to every position visited until time $\idxtimestep$. This smooth and dense reward encourages $\stageB(\cdot)$ to follow the visitation distributions predicted by $\stageA(\cdot)$.
The stop reward $\stopreward(\cdot)$ is only non-zero for $\stopaction$, when it is the earth mover's distance between the predicted goal distribution $\stopvisit_\idxtimestep$ and an empirical stopping distribution that assigns the full probability mass to the stop position in the policy execution. 
The exploration reward $\explorereward(\cdot)$ combines a positive reward for reducing the belief that the goal has not been observed yet (i.e., reducing $\stopvisit_\idxtimestep(\posoob)$) and a negative reward proportional to the probability that the goal position is unobserved according to $\stopvisit_\idxtimestep(\posoob)$. 
The action reward $\actionreward(\cdot)$ penalizes actions outside of the controller range.
Finally, $\stepweight$ is negative per-step reward to encourage efficient execution. We provide the reward implementation details in Appendix~\ref{seq:app:reward}.

Algorithm~\ref{algo:rl} shows the RL procedure. At every iteration, we collect $N$ simulation executions $\execposseq^{(1)}, ..., \execposseq^{(N)}$ using the policy $\stageB(\stageAsim(\cdot))$ (line \ref{ln:rl:exec}). 
To sample setpoint updates we treat the existing output as the mean of a normal distribution, add variance prediction, and sample the two velocities from the predicted normal distributions. 
We perform $K^{\rm RL}_{\rm steps}$ PPO updates using the return and value estimates (lines \ref{ln:rl:ppobegin}--\ref{ln:rl:ppoend}).
For every update, we sample state-action-return triplets from the collected trajectories, compute the PPO loss $\mathcal{L}_{PPO}(X,\valuefunc)$, update parameters $\paramsB$, and update the parameters $\valueparams$ of the value function $\valuefunc$. 
We pass the sampled executions to Process A to allow the model to learn to predict the visitation distributions in a way that is robust to the agent actions (line~\ref{ln:rl:mergeexec}).

\section{Experimental Setup}
\label{sec:experiment}

We provide the complete implementation and experimental setup details in Appendix~\ref{sec:app:exp}. 

\paragraph{Environment and Data} 
We use an Intel Aero quadcopter with a PX4 flight controller, and a Vicon motion capture system for pose estimates. 
For simulation, we use the quadcopter simulator from \citet{blukis2018following} that is based on Microsoft AirSim~\cite{airsim2017fsr}. 
The environment size is 4.7x4.7m. 
We randomly create environments with 5--8 landmarks, selected randomly out of a set of 15. 
Figure~\ref{fig:task} shows the real environment. 
We follow the crowdsourcing setup of \citet{misra2018mapping} to collect 997 paragraphs with 4,557 segments. We use 3,245/640/672 for training, development, and testing. 
We expand this data by concatenating consecutive segments to create more challenging instructions, including with exploration challenges~\cite{jain2019stay}. 
Figure~\ref{fig:task} shows an instruction made of two consecutive segments.
The simulator dataset $\dataset^\domainsim$ includes oracle demonstrations of all instructions, while the real-world dataset $\dataset^\domainreal$ includes only 402 single-segment demonstrations.
For evaluation in the physical environment, we sample 20 test paragraphs consisting of 93 single-segment and 73 two-segment instructions. 
We use both single and concatenated instructions for training, and test on each set separately. 
We also use the original \citet{misra2018mapping} data as additional simulation training data. 
Appendix~\ref{sec:app:data} provides data statistics and further details.

\paragraph{Evaluation}
We use human judgements to evaluate if the agent's final position is correct with regard to the instruction (goal score) and how well the agent followed the path described by the instruction (path score).
We present MTurk workers with an instruction and a top-down animation of the agent behavior, and ask for a 5-point Likert-scale score for the final position and trajectory correctness. We obtain five judgements per example per system. 
We also automatically measure (a) \textsc{SR}: success rate of stopping within 47cm of the correct position; and (b) \textsc{EMD}: earth mover's distance in meters between the agent and demonstration trajectories. 
Appendix~\ref{sec:app:eval} provides more evaluation details.

\paragraph{Systems}
We compare our approach, $\textsc{PVN2-\rail}$, with three non-learning and two learning baselines: 
(1) $\textsc{STOP}$: output the $\stopaction$ action without movement; 
(2) $\textsc{Average}$: take the average action for the average number of steps; 
(3) $\textsc{Oracle}$: a hand-crafted upper-bound expert policy that has access to the ground truth human demonstration; 
(4) $\textsc{PVN-BC}$ the \citet{blukis2018mapping} $\textsc{PVN}$ model trained with behavior cloning;
and (5) $\textsc{PVN2-BC}$: our model trained with behavior cloning. 
The two behavior cloning systems require access to an oracle that provides velocity command output labels, in addition to demonstration data. $\rail$ uses demonstrations, but does not require the oracle during learning. None of the learned systems use any oracle data at test time. All learned systems use the same training data $\dataset^{\domainsim}$ and $\dataset^{\domainreal}$, and include the domain-adaptation loss (Equation~\ref{eq:wass}).

\section{Results}
\label{sec:results}

\begin{figure*}[t]
\centering
\includegraphics[scale=0.24]{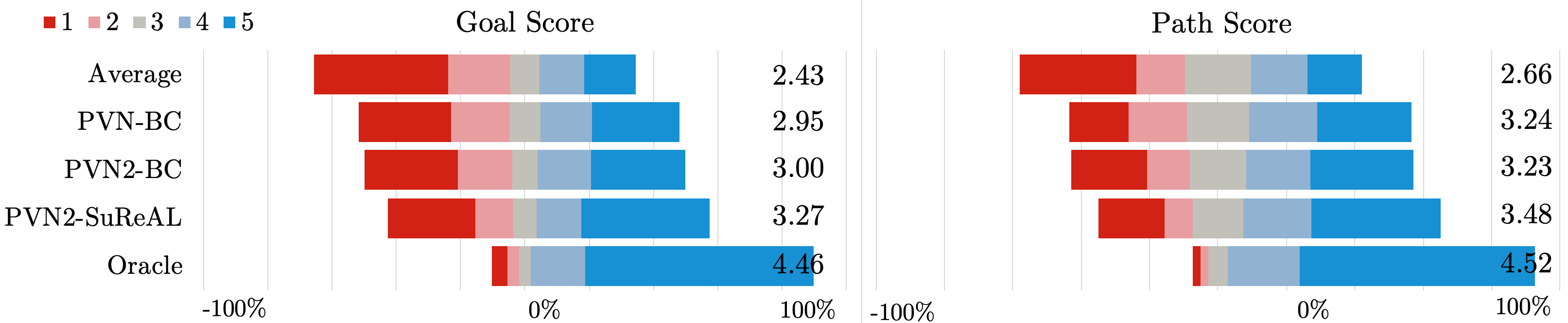}
\caption{Human evaluation results on the physical quadcopter on two-segment data. We plot the Likert scores using Gantt charts of score frequencies. The black numbers indicate average scores.}
\label{fig:human_results}
\end{figure*}

Figure~\ref{fig:human_results} shows human evaluation Likert scores.
Our model receives five-point scores 39.72\% of the time for getting the goal right, and 37.8\% of the time for the path. This is a 34.9\% and 24.8\% relative improvement compared to $\textsc{PVN2-BC}$, the next best system.
This demonstrates the  benefits of modeling observability, using $\rail$ for training-time exploration, and using a reward function that trades-off task performance and test-time exploration. 
The $\textsc{Average}$ baseline received only $15.8\%$ 5-point ratings in both \emph{path score} and \emph{goal score}, demonstrating the task difficulty.

\begin{figure}
\centering
\begin{minipage}{.61\textwidth}
\footnotesize
\begin{tabular}{|l|l|c|c|c|c|c|c|c|c|}
\hline
\multirow{2}{*}{} & \multirow{2}{*}{Method}   &\multicolumn{2}{|c|}{1-segment} & \multicolumn{2}{|c|}{2-segment}  \\
 \cline{3-10}
 &   & SR & EMD  & SR & EMD \\
\hline
\hline
\multicolumn{10}{|l|}{\textbf{Test Results}} \\
\hline
\multirow{5}{*}{\rotatebox[origin=c]{90}{Real}} 
& \avgmodel              & \phantom{0}37.0   & 0.42   & \phantom{0}16.7   & 0.71 \\
& \textsc{PVN-BC}        & \phantom{0}48.9   & 0.42   & \phantom{0}20.8   & 0.61 \\
& \textsc{PVN2-BC}      & \phantom{0}52.2   & 0.37   & \phantom{0}29.2   & 0.59 \\
& \textsc{PVN2-$\rail$} & \textbf{\phantom{0}56.5}   & \textbf{0.34}   & \textbf{\phantom{0}30.6}   & \textbf{0.52} \\
\ddline{2-10}
&\oraclemodel            & 100.0   & 0.17   & \phantom{0}91.7   & 0.23
 \\
\hline
\multirow{4}{*}{\rotatebox[origin=c]{90}{Sim}} 
& \avgmodel         & \phantom{0}29.5   & 0.53   & \phantom{0}8.7   & 0.80 \\
& \textsc{PVN-BC}   & \textbf{\phantom{0}64.1}   & 0.31   & \textbf{\phantom{0}37.5}   & 0.59 \\
& \textsc{PVN2-BC} & \phantom{0}55.4   & 0.34   & \phantom{0}34.7   & 0.58 \\
&  \textsc{PVN2-$\rail$} & \phantom{0}53.3   & \textbf{0.30}   & \phantom{0}33.3   & \textbf{0.42}\\
\ddline{2-10}
& \oraclemodel    &  100.0   & 0.13   & \phantom{0}98.6   & 0.17
 \\
\hline
\hline
\multicolumn{10}{|l|}{\textbf{Development Results}} \\
\hline
\multirow{4}{*}{\rotatebox[origin=c]{90}{Real}} 
& \textsc{PVN2-$\rail$}  & \phantom{0}54.8   & 0.32   & \phantom{0}31.0   & 0.50 \\
&\textsc{PVN2-$\rail$-noU} & \phantom{0}53.8   & 0.30   & \phantom{0}14.3   & 0.56 \\
&\textsc{PVN2-$\rail$}$_{50real}$  & \textbf{\phantom{0}60.6}   & \textbf{0.29}   & \textbf{\phantom{0}34.5}   & \textbf{0.44} \\
&\textsc{PVN2-$\rail$}$_{10real}$  & \phantom{0}46.2   & 0.33   & \phantom{0}17.9   & 0.56 \\
\hline
\multirow{2}{*}{\rotatebox[origin=c]{90}{Sim}} 
&\textsc{PVN2-$\rail$}  & \phantom{0}48.1   & 0.29   & \textbf{\phantom{0}39.3}   & \textbf{0.40} \\
& \textsc{PVN2-$\rail$-noU} & \phantom{0}53.8   & \textbf{0.28}   & \phantom{0}27.4   & 0.50 \\
& \textsc{PVN2-$\rail$-noI} & \textbf{\phantom{0}56.2}   & \textbf{0.28}   & \phantom{0}25.9   & 0.45 \\
\hline
\end{tabular}
\captionof{table}{Automated evaluation test and development results.  SR: \emph{success rate} (\%) and EMD: \emph{earth-mover's distance} in meters between agent and demonstration trajectories.}
\label{tab:auto_results}
\end{minipage}~~
\begin{minipage}{.38\textwidth}
  \centering
  \vspace{0pt}
  \includegraphics[scale=0.27]{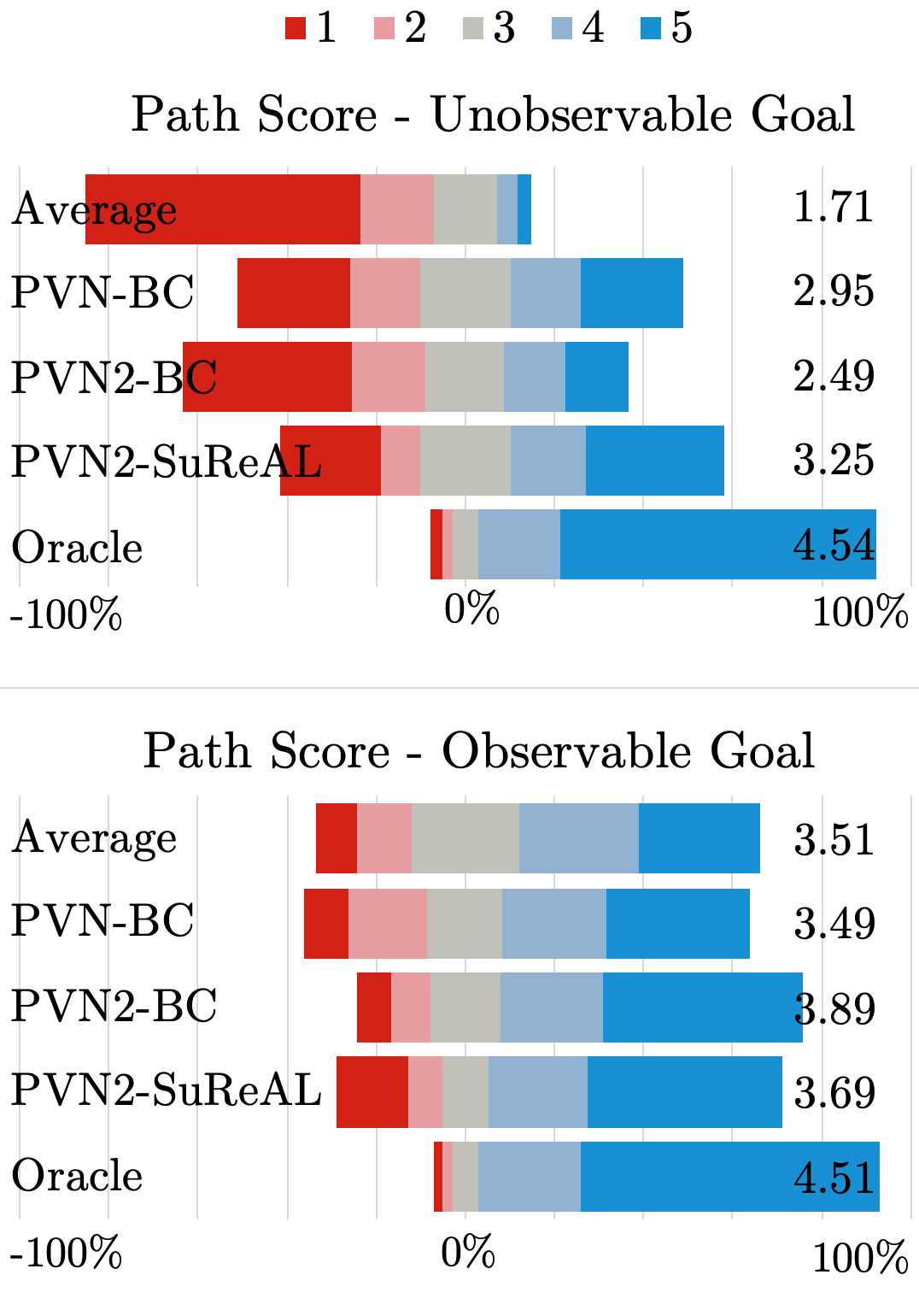}
  \captionof{figure}{Human evaluation path score frequencies (Figure~\ref{fig:human_results}) decomposed by initially unobservable (top) or observable (bottom) goal location.}
  \label{fig:human_results_obs}
\end{minipage}
\end{figure}

We study how well our model addresses observability and exploration challenges. 
Figure~\ref{fig:human_results_obs} shows human path score judgements split to tasks where the goal is visible from the agent's first-person view at start position (34 examples) and those where it is not and exploration is required (38 examples).  
Our approach outperforms the baselines in cases requiring exploration, but it is slightly outperformed by $\textsc{PVN2-BC}$ in simple examples. This could be explained by our agent attempting to explore the environment in cases where it is not necessary.

Table~\ref{tab:auto_results} shows the automated metrics for both environments. 
We observe that the success rate (SR) measure is sensitive to the threshold selection, and correct executions are often considered as wrong; $\textsc{PVN2-\rail}$ gets 30.6\% SR  compared to  a perfect human score 39.72\% of the time.
This highlights the need for human evaluation, and must be considered when interpreting the SR results. 
We generally find \textsc{EMD} more reliable, although it also does not account for semantic correctness. 

Comparing to $\textsc{PVN2-BC}$, our approach performs better on the real environment demonstrating the benefit of $\rail$. In simulation, we observe better $\textsc{EMD}$, but worse SR. Qualitatively, we observe our approach often recovers the correct overall trajectory, with a slightly imprecise stopping location due to instruction ambiguity or control challenges. Such partial correctness is not captured by SR.
Comparing $\textsc{PVN2-BC}$ and $\textsc{PVN-BC}$, we see the benefit of modeling  observability. $\rail$ further improves upon $\textsc{PVN2-BC}$, by learning to explore unobserved locations at test-time.

Comparing our approach between simulated and real environments, we see an absolute performance degradation of 2.7\% SR and 0.1 EMD from simulation to the real environment. 
This highlights the remaining challenges of visual domain transfer and complex flight dynamics. 
The flight dynamics challenges are also visible in the  $\oraclemodel$ performance degradation between the two environments.

We study several ablations. First, we quantify the effect of using a smaller number of real-world training demonstrations. We randomly select subsets of demonstrations, with the constraint that all objects are visually represented. We find that using only half (200) of the physical demonstrations does not appear to reduce performance ($\textsc{PVN2-\rail}_{50real}$), while using only 10\% (40), drastically hurts real-world performance ($\textsc{PVN2-\rail}_{10real}$). This shows that the learning method is successfully leveraging real-world data to improve performance, while requiring relatively modest amounts of data.
We also study performance without access to the instruction ($\textsc{PVN2-\rail-noU}$), and with using a blank input image ($\textsc{PVN2-\rail-noI}$). 
The relatively high SR of these ablations on 1-segment instructions highlights the inherent bias in simple trajectories. The 2-segment data, which is our main focus, is much more robust to such biases. %
Appendix~\ref{sec:app:results} provides more  automatic evaluation results, including additional ablations and results on the original data of \citet{misra2018mapping}.

\vspace{-3pt}
\section{Discussion}
\label{sec:conclusion}
\vspace{-3pt}

We study the problem of mapping natural language instructions to continuous control of a physical quadcopter drone.
Our two-stage model decomposition allows some level of re-use and modularity. For example, a trained Stage 1 can be re-used with different robot types. 
This decomposition and the interpretability it enables also create limitations, including limited sensory input for deciding about control actions given the visitation distributions. 
These are both important topics for future study.

Our learning method, $\rail$, uses both annotated demonstration trajectories and a reward function. 
In this work, we assume demonstration trajectories  were generated with an expert policy. 
However, $\rail$ does not necessarily require the initial demonstrations to come from a reasonable policy, as long as we have access to the gold visitation distributions, which are easier to get compared to oracle actions. 
For example, given an initial policy that immediately stops instead of demonstrations, we will train Stage 1 to predict the given visitation distributions and Stage 2 using the intrinsic reward. 
Studying this empirically is an important direction for future work. 

Finally, our environment provides a strong testbed for our system-building effort and the transition from the simulation to the real world. However, various problems are not well represented, such as reasoning about obstacles, raising important directions for future work. While we do not require the simulation to accurately reflect the real world, studying scenarios with stronger difference between the two is another important future direction.
Our work also points towards the need for better automatic evaluation for instruction following, or, alternatively, wide adoption of human evaluation. 

\acknowledgments{This research was supported by the generosity of Eric and Wendy Schmidt by recommendation of the Schmidt Futures program, a Google Faculty Award, NSF CAREER-1750499, AFOSR FA9550-17-1-0109, an Amazon Research Award, and cloud computing credits from Amazon. We thank Dipendra Misra, Alane Suhr, and the anonymous reviewers for their helpful comments.}

{
\setlength{\bibsep}{3pt}
\bibliography{references}
}

\clearpage

\appendix

\section{Execution Examples on Real Quadcopter}
\begin{figure}[t]
\centering
\includegraphics[scale=0.32]{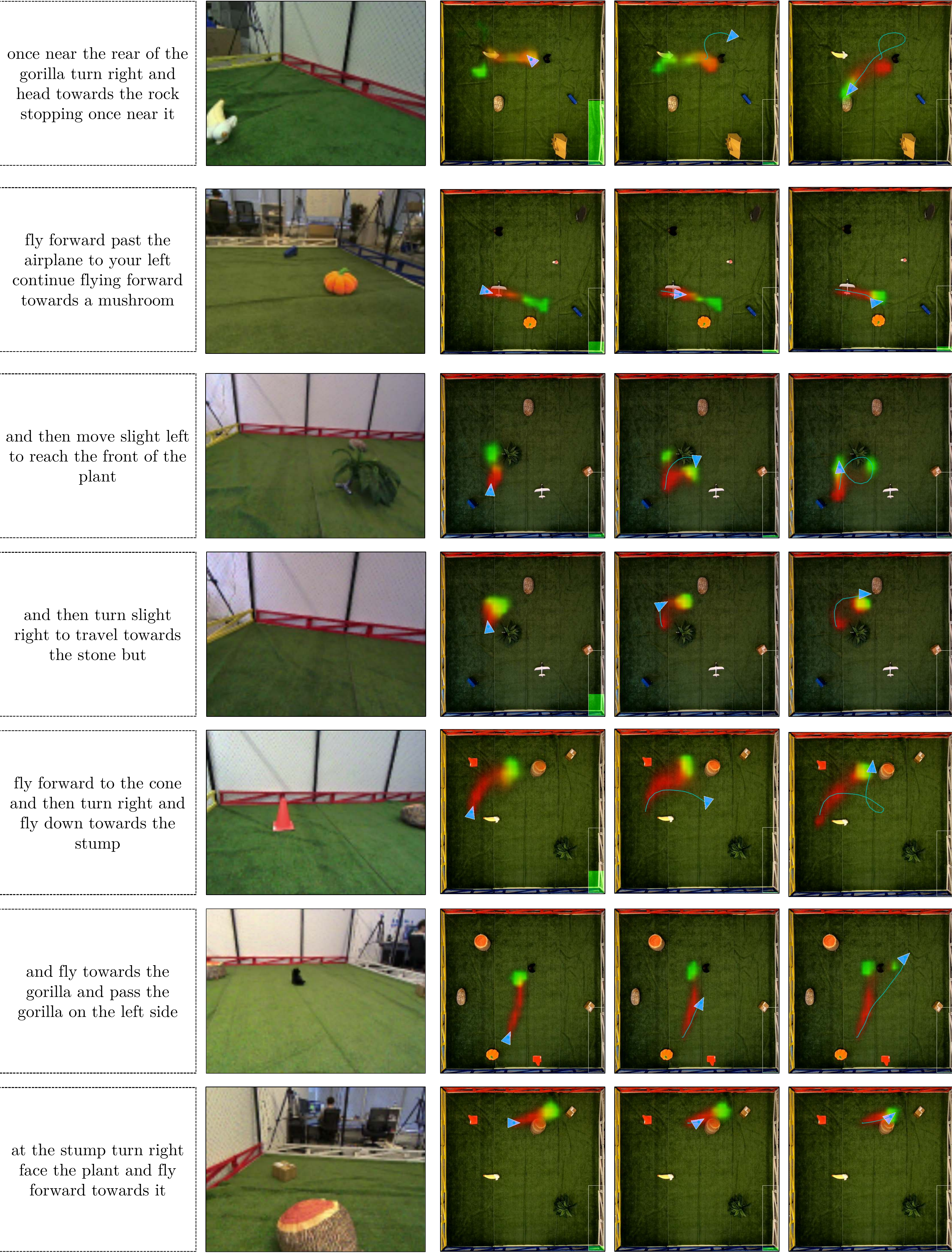}
\caption{Instruction executions from the development set on the physical quadcopter. For each example, the figure shows (from the left) the input instruction, the initial image that the agent observes, the initial visitation distributions overlaid on the top-down view, visitation distributions at the midpoint of the trajectory, and the final visitation distributions when outputting the $\stopaction$ action. The green bar on the lower-right corner of each distribution plot shows the predicted probability that the goal is not yet observed. The blue arrow indicates the agent pose.
}
\label{fig:exec_examples}
\end{figure}

\paragraph{Examples of Different Instruction Executions}
Figure~\ref{fig:exec_examples} shows a number of instruction-following executions collected on the real drone, showing successes and some typical failures.

\begin{figure}[t]
\centering
\includegraphics[scale=0.32]{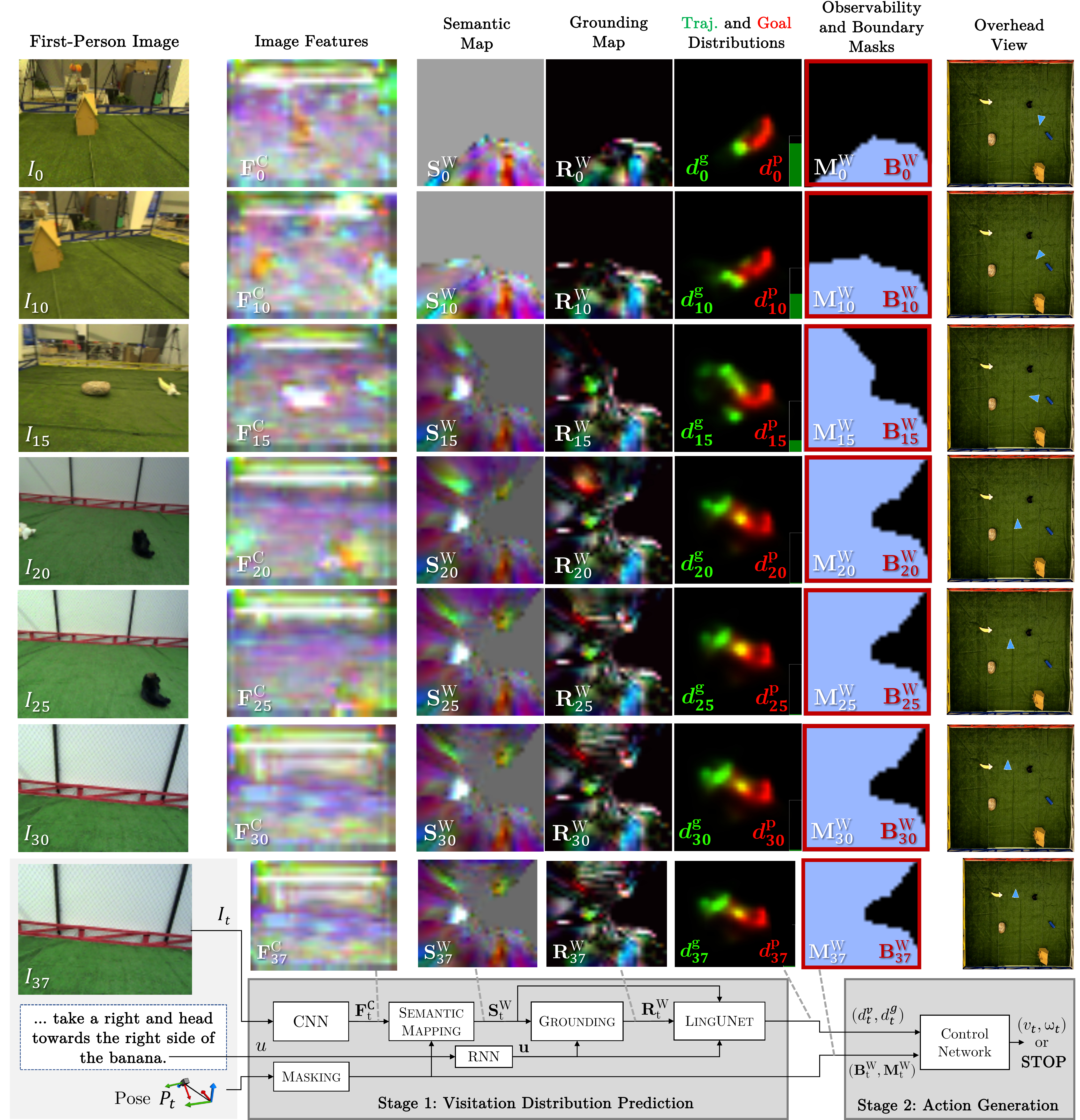}
\caption{
Illustration of changes in the intermediate representations during an instruction execution, showing how information is accumulated in the semantic maps over time, and how that affects the predicted visitation distributions. We show the instruction from Figures~\ref{fig:task}-\ref{fig:model}.  From top to bottom, representations are shown for timesteps 0, 10, 15, 20, 25, 30, and 37 (the final timestep). From left to right, we show the input image $\image_t$, first-person features $\fmcam_t$, semantic map $\smworld_t$, grounding map $\rmworld_t$, goal and position visitation distributions $\stopvisit_t$ and $\trajvisit_t$, observability mask $\maskworld_t$ and boundary mask $\boundaryworld_t$, and the  overhead view of the environment. The agent position is indicated with a blue arrow in the overhead view. The agent does not have access to the overhead view, which is provided for illustration purposes only.}
\label{fig:exec_repr}
\vspace{10pt}
\end{figure}

\paragraph{Visualization of Intermediate Representations}
Figure~\ref{fig:exec_repr} shows the intermediate representations and visitation predictions over time during instruction execution for the examples used in Figures~\ref{fig:task}-\ref{fig:model}, illustrating the model reasoning. The model is able to output the $\stopaction$ action to stop on the right side of the banana, even after the banana has disappeared from the first-person view. This demonstrates the advantages of using an explicit spatial map aggregated over time instead, for example, a learned hidden state vector representing the agent state.

\section{Model Details}
\label{app:sec:model}

\subsection{Stage 1: Visitation Distribution Prediction}
\label{app:sec:model:stage1}

The first-stage model is largely based on the Position Visitation Network, except for several improvements we introduce:
\begin{itemize}
    \item Computing observability and boundary masks $\maskworld$ and $\boundaryworld$ that are used to track unexplored space and environment boundaries.
    \item Introducing a placeholder position $\posoob$ that represents all unobserved positions in the environment for use in the visitation distributions.
    \item Modification to the $\lingunet$ architecture to support outputting a probability score for the unobserved placeholder position, in addition to the 2D distributions over environment positions.
    \item Predicting 2D probability distributions only over observed locations in the environment.
    \item Minor hyperparameter changes to better support the longer instructions.
\end{itemize}

\begin{center}
\fbox{\begin{minipage}{0.95\linewidth}\textbf{\emph{The description in Section~\ref{app:sec:model:stage1} has been taken  from \citet{blukis2018mapping}. We present it here for reference and completeness, with minor modifications to highlight technical differences.}}\end{minipage}}
\end{center}

\subsubsection{Instruction Representation}

We represent the instruction $\instruction = \langle \instruction_1, \cdots \instruction_{\instrlen} \rangle$ as an embedded vector $\instructionemb$. We generate a series of hidden states $\mathbf{h}_i = \textsc{LSTM}(\phi(\instruction_i), \mathbf{h}_{i-1})$, $i = 1\dots \instrlen$, where $\textsc{LSTM}$ is a Long-Short Term Memory recurrent neural network (RNN) and $\phi$ is a learned word-embedding function. The instruction embedding is  the last hidden state $\instructionemb = \mathbf{h}_{\instrlen}$.
This part is replicated as is from \citet{blukis2018mapping}.

\subsubsection{Semantic Mapping}
\begin{figure}[t]
\centering
\includegraphics[scale=0.3]{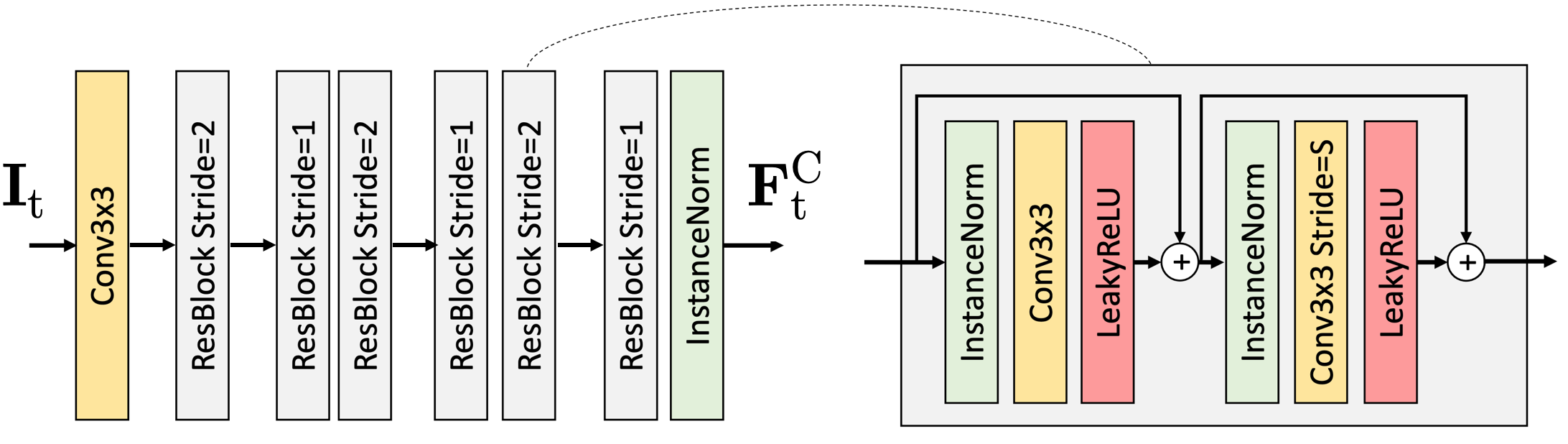}
\caption{The 13-layer ResNet architecture used in PVN and PVN2 networks (figure adapted from~\citet{blukis2018following}).}
\label{fig:resnet}
\end{figure}

We construct the semantic map using the method of \citet{blukis2018following}.
The full details of the process are specified in the original paper.
Roughly speaking, the semantic mapping process includes three steps: feature extraction, projection, and accumulation.
At timestep $\idxtimestep$, we process the currently observed image $\image_\idxtimestep$ using a 13-layer residual neural network $\resnet$ (Figure~\ref{fig:resnet}) to generate a feature map $\fmcam_\idxtimestep = \resnet(\image_\idxtimestep)$ of size $W_f \times H_f \times C$.
We compute a feature map in the world coordinate frame $\fmworld_\idxtimestep$ by projecting $\fmcam_\idxtimestep$ with a pinhole camera model onto the ground plane at elevation zero.

The semantic map of the environment $\smworld_\idxtimestep$ at time $t$ is an integration of $\fmworld_\idxtimestep$ and $\smworld_{\idxtimestep-1}$, the  map from the previous timestep.
The integration equation is given in Section 4c in \citet{blukis2018following}.
This process generates a tensor  $\smworld_{\idxtimestep}$  of size $W_w \times H_w \times C$ that represents a map, where each location $[\smworld_{\idxtimestep}]_{(x,y)}$ is a $C$-dimensional feature vector computed from all past observations $\image_{<\idxtimestep}$, each processed to learned features $\fmcam_{<\idxtimestep}$ and  projected onto the environment ground in the world frame at coordinates $(x,y)$.
This map maintains a learned high-level representation for every world location $(x,y)$ that has been visible in any of the previously observed images.
We define the world coordinate frame using the agent starting pose $\pose_1$; the agent start position is the coordinates $(0,0)$, and the positive direction of the $x$-axis is along the agent heading.
This gives consistent meaning to spatial language, such as \nlstring{turn left} or \nlstring{pass on the left side of}.

\subsubsection{Grounding}
We create the \emph{grounding map} $\rmworld_t$ with a 1$\times$1 convolution $\rmworld_\idxtimestep = \smworld_\idxtimestep \circledast \kernel_G$. The kernel $\kernel_G$ is computed using a learned linear transformation  $\kernel_G = \weights_G \instructionemb + \bias_G$, where $\instructionemb$ is the instruction embedding.
The grounding map $\rmworld_\idxtimestep$ has the same height and width as $\smworld_\idxtimestep$, and during training we optimize the parameters so it captures the objects mentioned in the instruction $\instruction$ (Section~\ref{sec:app:aux}).

\subsubsection{$\lingunet$ and Visitation Distributions}

\begin{figure}[t]
\centering
\includegraphics[scale=0.4]{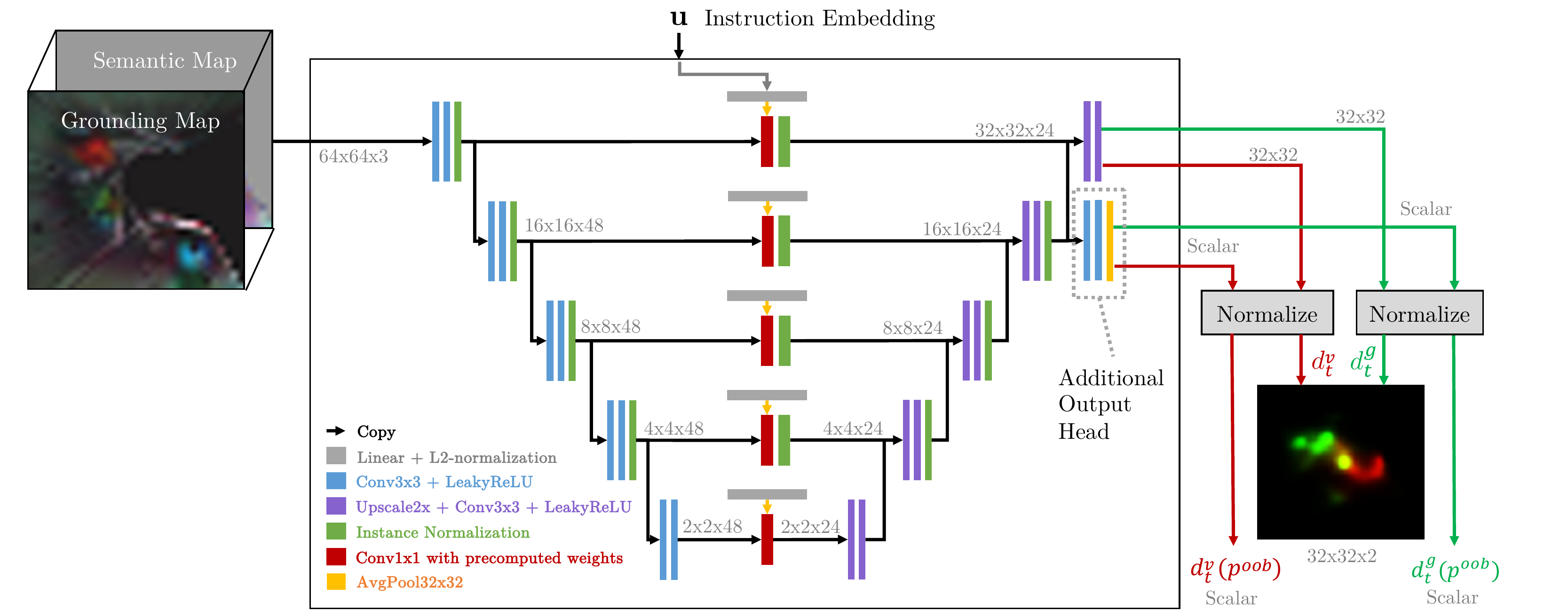}
\caption{The $\lingunet$ architecture, showing the additional output head that was added as part of the PVN2 model. $\lingunet$ outputs raw scores, which we normalize over the domain of each distribution. This figure is adapted from~\citet{blukis2018mapping}.}
\label{fig:lingunet}
\vspace{10pt}
\end{figure}

The following paragraphs are adapted from \citet{blukis2018mapping} and formally define the $\lingunet$ architecture with our modifications.  Figure~\ref{fig:lingunet} illustrates the architecture.

$\lingunet$ uses a series of convolution and scaling operations.
The input map $\featmap_0 = [\smworld_\idxtimestep, \rmworld_\idxtimestep]$ is processed through $L$ cascaded convolutional layers to generate a sequence of feature maps $\featmap_k = \conv^{D}_k(\featmap_{k-1})$, $k = 1\dots L$.\footnote{$[\cdot,\cdot]$ denotes concatenation along the channel dimension.}
Each  $\featmap_k$ is filtered with a 1$\times$1 convolution with weights $\kernel_k$. The kernels $\kernel_k$ are computed from the instruction embedding $\instructionemb$ using a learned linear transformation $\kernel_k = \weights^{u}_{k} \instructionemb + \bias^{u}_{k}$.
This generates $l$ language-conditioned feature maps $\featmaptxt_k = \featmap_k \circledast \kernel_k$, $k = 1\dots L$.
A series of $L$ upscale and convolution operations computes $L$ feature maps of increasing size:

\begin{small}
\begin{eqnarray*}
\featmapdeconv_k = \left\{\begin{array}{lr}
        {\upscale(\conv^U_k([\featmapdeconv_{k+1}, \featmaptxt_{k}]))}, & \text{if } 1 \leq k \leq L-1\\
        {\upscale(\conv^U_k(\featmaptxt_{k}))}, & \text{if } k=L
        \end{array}\right. \;\;,
\end{eqnarray*}
\end{small}

We modify the original $\lingunet$ design by adding an output head that outputs a vector $\lingunetvecout$:

\begin{small}
\begin{equation}
    \lingunetvecout = \avgpool(\conv^\lingunetvecout(\featmapdeconv_2))\;\;,\nonumber
\end{equation}
\end{small}

where $\avgpool$ takes averages across the  dimensions.

The output of $\lingunet$ is a tuple ($\featmapdeconv_1$, $\lingunetvecout$), where $\featmapdeconv_1$ is of size $W_w \times H_w \times 2$ and $\lingunetvecout$ is a vector of length 2. This output is used to compute two distributions, and can be increased if more distribution are predicted, such as in \citet{Suhr2019:cerealbar}. We use an additional normalization step to produce the position visitation and goal visitation distributions given ($\featmapdeconv_1$, $\lingunetvecout$).

\subsection{Control Network: Action Generation and Value Function}
\label{sec:app:control}

\begin{figure}[t]
\centering
\includegraphics[scale=0.38]{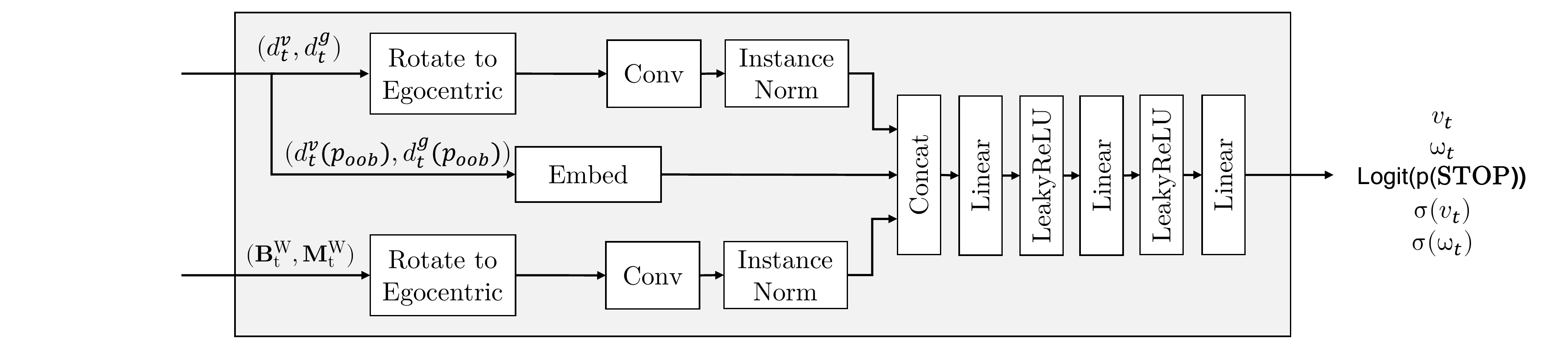}
\caption{Control network architecture.}
\label{fig:app:control}
\end{figure}

Figure~\ref{fig:app:control} shows the architecture of the control network for the second action generation stage of the model. The value function architecture is identical to the action generator and also uses the control network, except that it has only a single output. The value function does not share the parameters with the action generator.

The control network takes as input the trajectory and stopping visitation distributions $\trajvisit_\idxtimestep$ and $\stopvisit_\idxtimestep$, as well as the observability and boundary masks $\maskworld_\idxtimestep$ and $\boundaryworld_\idxtimestep$.
The distributions $\trajvisit_\idxtimestep$ and $\stopvisit_\idxtimestep$ are represented as 2D square-shaped images over environment locations, where unobserved locations have a probability of zero. Additional scalars $\trajvisit(\posoob)$ and $\stopvisit(\posoob)$ define the probability mass outside of any observed environment location.

The visitation distributions $\trajvisit_\idxtimestep$ and $\stopvisit_\idxtimestep$ are first rotated to the agent's current ego-centric reference frame, concatenated along the channel dimension, and then processed with a convolutional neural network. The output is flattened into a vector.
The masks $\boundaryworld_\idxtimestep$ and $\maskworld_\idxtimestep$ are processed in an analogous way to the visitation distributions $\trajvisit_\idxtimestep$ and $\stopvisit_\idxtimestep$, and the output is also flattened into a vector.
The scalars $\trajvisit(\posoob)$ and $\stopvisit(\posoob)$ are embedded into fixed-length vectors:

\begin{small}
\begin{eqnarray}
    \textsc{embed}_{\stopvisit(\posoob)} &=& \textbf{q}_{1} \cdot \stopvisit(\posoob) - \textbf{q}_{1} \cdot (1 - \stopvisit(\posoob)) \nonumber \\
    \textsc{embed}_{\trajvisit(\posoob)} &=& \textbf{q}_{2} \cdot \trajvisit(\posoob) - \textbf{q}_{2} \cdot (1 - \trajvisit(\posoob))\;\;, \nonumber
\end{eqnarray}
\end{small}

where $\textbf{q}_{(\cdot)}$ are random fixed-length vectors. We do not tune $\textbf{q}_{(\cdot)}$.

The resulting vector representations for the visitation distributions, out-of-bounds probabilities, and masks are concatenated and processed with a three-layer multi-layer perceptron (MLP). The output are five scalars. Two of the scalars are predicted forward and angular velocities $v_t$ and $\omega_t$, one scalar is the logit of the stopping probability, and two scalars are standard deviations used during PPO training to define a Gaussian probability distribution over actions.

\subsection{Coordinate Frames of Reference}
At the start of each task, we define the world reference frame according to the agent's starting position, with x and y axis pointing forward and left respectively, according to the agent's position. The maps are represented with the origin at the center. Throughout the instruction execution, this reference frame remains fixed.
Within the first model stage, the semantic and grounding maps, observability and boundary masks, and visitation distributions are all represented in the world reference frame.
At the input to second stage, we transform the visitation distributions, and observability and boundary masks to the agent's current egocentric frame of reference. This allows the model to focus on generating velocities to follow the high probability regions, without having to reason about coordinate transformations.

\section{Additional Learning Details}

\subsection{Discriminator Architecture and Training}
\label{app:discriminator}

\begin{figure}[t]
\vspace{20pt}
\centering
\includegraphics[scale=0.5]{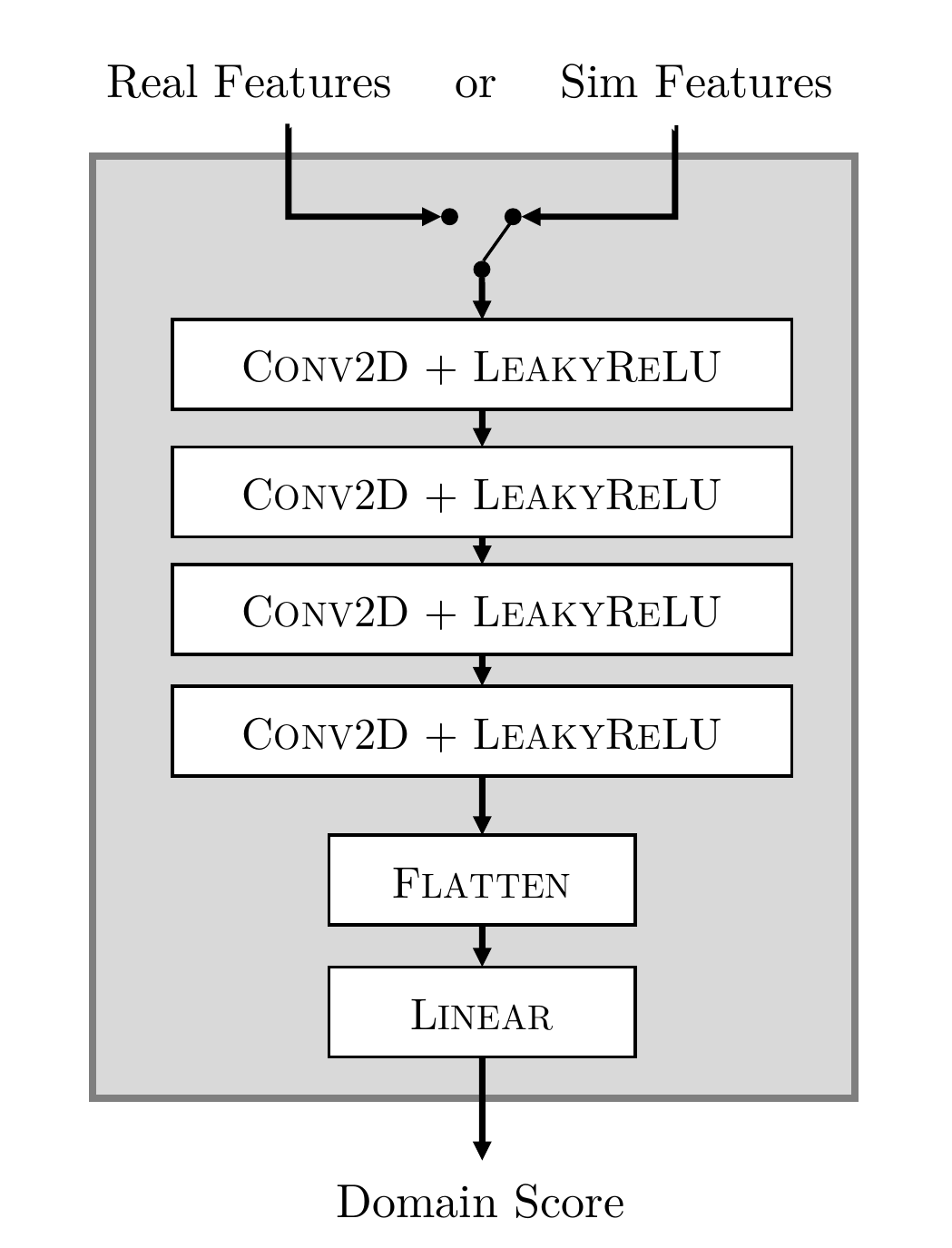}
\caption{Our discriminator architecture. The discriminator takes as input a 3D feature map with two spatial dimensions and one feature dimension. It processes the feature map with a cascade of four convolutional neural networks with LeakyReLU non-linearities, before processing the output with a linear neural network layer. The discriminator is trained to output a scalar score that assigns high values to feature maps from the simulated domain, and low values from the real domain. The discriminator is used as component in our Wasserstein domain loss $\mathcal{L}_{W}$.}
\label{fig:discriminator}
\end{figure}

Figure~\ref{fig:discriminator} shows the neural network architecture of our discriminator $\discriminator$.
The Wasserstein distance estimation procedure from~\citet{shen2017wasserstein} requires a discriminator that is K-Lipschitz continuous. We guarantee that our discriminator meets this requirement by clamping the discriminator parameters $\discrimparams$ to a range of $[-\clamp; \clamp]$ after every gradient update~\cite{shen2017wasserstein}.

\subsection{Return Definition}
The expected return $R_t(\execposseq)$ is defined as:

\begin{small}
\begin{equation}
    R_{t}(\execposseq) = \sum_{\substack{i \geq t, (\state_i,\action_i)\in \execposseq,  \\ \context_{i} = \gencontexts(\state_i)}}\gamma^{i-t}\reward(\context_{i},\action_{i})\;\;, \nonumber
\label{eq:return}
\end{equation}
\end{small}

where $\execposseq$ is a policy execution, $\gencontexts(\state_i)$ is the agent context observed at state $\state_i$, $\gamma$ is a discount factor, and $\reward(\cdot)$ is the intrinsic reward. The reward does not depend on any external state information, but only on the agent context and visitation predictions.

\subsection{Reward Function}
\label{seq:app:reward}

Section~\ref{sec:learning} provides the high level description and motivation of the intrinsic reward function.

\paragraph{Visitation Reward}
We  design the visitation reward to reward policy executions $\execposseq$ that closely match the predicted visitation distribution $\trajvisit$. An obvious choice would be the probability of the trajectory under independence assumptions  $P(\execposseq) \approx  \prod_{\position \in \execposseq} \trajvisit_\idxtimestep(\position)$.
According to this measurement, if $\trajvisit_\idxtimestep(\position) = 0$ for any $\position \in \execposseq$, then $P(\execposseq) = 0$. This would lead to a reward of zero as soon as the policy makes a mistake, resulting in sparse rewards and slow learning. Instead, we define the visitation reward in terms of earth mover's distance that provides a smooth and dense reward. The visitation reward $\visitreward$ is:

\begin{small}
\begin{equation}
    \visitreward(\context_t, \action_t) = \visitpotential(\context_{t}, \action_{t}) - \visitpotential(\context_{t-1}, \action_{t-1})\;\;, \nonumber %
\end{equation}
\end{small}

where $\visitpotential$ is a reward shaping  potential:

\begin{small}
\begin{equation}
    \visitpotential(\context_{\idxtimestep}, \action_{\idxtimestep}) =     -\emd(\mathbbm{1}_{\position \in \posseq},     \trajvisit_\idxtimestep(\position_\idxtimestep \mid \position_\idxtimestep \in \observedpositions))\;\;. \nonumber
\end{equation}
\end{small}

$\emd$ is the earth mover's distance in Euclidean $\reals^2$ space, $\mathbbm{1}_{\position \in \posseq}$ is a probability distribution that assigns equal probability to all positions visited thus far, $\observedpositions$ is the set of positions the agent has observed so far,\footnote{We restrict the metric to observed locations on the map, because as discussed in Section \ref{sec:model}, all unobserved locations are represented by a dummy location $\posoob \notin \mathbbm{R}^2$.} and $\trajvisit_\idxtimestep(\position_\idxtimestep \mid \position_\idxtimestep \in \observedpositions)$ is the position visitation distribution over all observed positions.
Intuitively, $\visitreward$ rewards per-timestep reduction in earth mover's distance between the predicted visitation distribution and the empirical distribution derived from the agent's trajectory.

\paragraph{Stop Reward}
Similar to $\visitreward$, the stop reward $\stopreward$ is the negative earth mover's distance between the conditional predicted goal distribution over all observed environment locations, and the empirical stop distribution $\mathbbm{1}_{\position = \execposseq_{-1}}$ that assigns unit probability to the agent's final stopping position.

\begin{small}
\begin{equation}
    \stopreward(\context_\idxtimestep, \action_\idxtimestep) = - \mathbbm{1}_{\action_\idxtimestep = \stopaction} \cdot \emd(\mathbbm{1}_{\position = \execposseq_{-1}}, \stopvisit_\idxtimestep(\position_\idxtimestep \mid \position_\idxtimestep \in \observedpositions))\;\;. \nonumber
\end{equation}
\end{small}

\paragraph{Exploration Reward}
The exploration reward $\explorereward$ is:

\begin{small}
\begin{equation}
    \explorereward(\context_\idxtimestep, \action_\idxtimestep) =
    (\explorepotential(\context_\idxtimestep, \action_\idxtimestep) -
    \explorepotential(\context_{\idxtimestep-1}, \action_{\idxtimestep-1})) - \mathbbm{1}_{\action_\idxtimestep = \stopaction} \cdot \stopvisit_\idxtimestep(\posoob)\;\;, \label{eq:explorereward}
\end{equation}
\end{small}

where:

\begin{small}
\begin{equation}
    \explorepotential(\context_\idxtimestep, \action_\idxtimestep) = \max_{t'<\idxtimestep}[1 - \stopvisit_{t'}(\posoob)]\;\;. \nonumber
\end{equation}
\end{small}

The term $\explorepotential$ reflects the agent's belief that it has observed the goal location $\goalpos$. $1 - \stopvisit_{t'}(\posoob)$ is the probability that the goal has been observed before time $t'$. We take the maximum over past timesteps to reduce effects of noise from the model output. The second term in Equation~$\ref{eq:explorereward}$ penalizes the agent for stopping while it predicts that the goal is not yet observed.

\subsection{Auxiliary Objectives}
\label{sec:app:aux}

During training, we add an additional auxiliary loss $\mathcal{L}_{aux}$ to the supervised learning loss $\mathcal{L}_{SL}$ to ensure that the different modules in the PVN model specialize according to their function. The auxiliary loss is:

\begin{small}
\begin{equation}
\mathcal{L}_{aux}(\context_t) = \mathcal{L}_{\rm percept}(\context_t) + \mathcal{L}_{\rm ground}(\context_t) + \mathcal{L}_{\rm lang}(\context_t)\;\;.
\end{equation}
\end{small}

\begin{center}
\fbox{\begin{minipage}{0.95\linewidth}\textbf{\emph{The text in the remainder of Section~\ref{sec:app:aux} has been taken from~\citet{blukis2018mapping}. We  present it here for reference and completeness.}}\end{minipage}}
\end{center}

\paragraph{Object Recognition Loss}

The object-recognition loss $\mathcal{L}_{\rm percept}$ ensures  the semantic map $\smworld_\idxtimestep$ stores information about locations and identities of objects.
At timestep $\idxtimestep$, for every object $o$ that is visible in the first person image $\image_\idxtimestep$, we classify the feature vector in the position in the semantic map $\smworld_\idxtimestep$ corresponding to the object location in the world. We use a linear softmax classifier to predict the object identity given the feature vector. At a given timestep $\idxtimestep$ the classifier loss is:

\begin{small}
\begin{equation*}
\mathcal{L}_{\rm percept}(\params_1) = \frac{-1}{|O_{{\rm FPV}}|}\sum_{o \in O_{{\rm FPV}}}[\hat{y}_o log(y_o)]\;\;,
\end{equation*}
\end{small}

where $\hat{y}_o$ is the true class label of the object $o$ and $y_o$ is the predicted probability. $O_{{\rm FPV}}$ is the set of objects visible in the image $\image_\idxtimestep$.

\paragraph{Grounding Loss}

For every object $o$ visible in the first-person image $\image_\idxtimestep$, we use the feature vector from the grounding map $\rmworld_\idxtimestep$ corresponding to the object location in the world with a linear softmax classifier to predict whether the object was mentioned in the instruction $u$. The objective is:

\begin{small}
\begin{equation*}
\mathcal{L}_{\rm ground}(\params_1) = \frac{-1}{|O_{{\rm FPV}}|}\sum_{o \in O_{{\rm FPV}}}[\hat{y}_o log(y_o) + (1-\hat{y}_o) log(1 - y_o)]\;\;,
\end{equation*}
\end{small}

where $\hat{y}_o$ is a 0/1-valued label indicating whether the object o was mentioned in the instruction and $y_o$ is the corresponding model prediction. $O_{{\rm FPV}}$ is the set of objects visible in the image $\image_\idxtimestep$.

\paragraph{Language Loss}

The instruction-mention auxiliary objective uses a similar classifier to the grounding loss. Given  the instruction embedding $\instructionemb$, we predict for each of the possible objects whether it was mentioned in the instruction $u$. The objective is:

\begin{small}
\begin{equation}
\nonumber \mathcal{L}_{\rm lang}(\params_1) = \frac{-1}{|O|}\sum_{o \in O_{{\rm FPV}}}[\hat{y}_o log(y_o) + (1-\hat{y}_o) log(1 - y_o)]\;\;,
\end{equation}
\end{small}

where $\hat{y}_o$ is a 0/1-valued label, same as above.

\paragraph{Automatic Word-object Alignment Extraction}

In order to infer whether an object $o$ was mentioned in the instruction $u$, we use automatically extracted word-object alignments from the dataset. Let $E(o)$ be the event that an object $o$ occurs within 15 meters of the human-demonstration trajectory ${\posseq}$, let $E(\tau)$ be the event that a word type $\tau$ occurs in the instruction $\instruction$, and let $E(o, \tau)$ be the event that both $E(o)$ and $E(\tau)$ occur simultaneously. The pointwise mutual information between events $E(o)$ and $E(\tau)$ over the training set is:

\begin{small}
\begin{equation*}
  \textsc{PMI}(o, \tau) = P(E(o,\tau)) \log \frac{P(E(o,\tau))}{P(E(o))P(E(\tau))}\;\;,
\end{equation*}
\end{small}

where the probabilities are estimated from counts over training examples $\{ (\instruction^{(i)}, \state_1^{(i)},  \posseq^{(i)})\}_{i = 1}^N$.
The output set of word-object alignments is:

\begin{small}
\begin{equation*}
  \{(o, \tau) \: | \: \textsc{PMI}(o, \tau) > T_{\rm PMI} \wedge P(\tau) < T_{\tau}\}\;\;,
\end{equation*}
\end{small}

where $T_{PMI} = 0.008$ and $T_{\tau} = 0.1$ are threshold hyperparameters.

\section{Dataset Details}
\label{sec:app:data}

\begin{table}[t]
\begin{center}
\footnotesize
\centering
\begin{tabular}{|l|l|l|c|c|c|c|c|c|}
\hline
Dataset & Type & Split & \# Paragraphs & \# Instr. & Avg. Instr. Len. (tokens) & Avg. Path Len. (m) \\
\hline

\multirow{6}{*}{\textsc{Lani}} &
\multirow{3}{*}{1-segment} &
Train & 4200  &  19762  &  11.04  &  1.53\\
& & Dev & 898  &  4182  &  10.94  &  1.54\\
& & Test &  902  &  4260  &  11.23  &  1.53\\
\ddline{2-7}
& \multirow{3}{*}{2-segment} &
Train & 4200  &  15919  &  21.84  &  3.07\\
& & Dev & 898  &  3366  &  21.65  &  3.10\\
& & Test & 902  &  3432  &  22.26  &  3.07\\
\hline
\multirow{6}{*}{\textsc{Real}} &
\multirow{3}{*}{1-segment} &
Train & 698  &  3245  &  11.10  &  1.00\\
& & Dev & 150  &  640  &  11.47  &  1.06\\
& & Test &  149  &  672  &  11.31  &  1.06\\
\ddline{2-7}
& \multirow{3}{*}{2-segment} &
Train & 698  &  2582  &  20.97  &  1.91\\
& & Dev & 150  &  501  &  21.42  &  2.01\\
& & Test &  149  &  531  &  21.28  &  1.99\\
\hline
\end{tabular}
\caption{Dataset and split sizes. \textsc{Lani} was introduced by \citet{misra2018mapping}. Each layout in \textsc{Lani} consists of 6--13 landmarks out of a total of 64. \textsc{Real} is the additional data we collected for use on the physical drone. In \textsc{Real}, each layout has 5--8 landmarks from a set of 15 that is a subset of the landmarks in \textsc{Lani}.}
\label{tab:dataset_sizes}
\end{center}
\end{table}

\paragraph{Natural Language and Demonstration Data}
Table~\ref{tab:dataset_sizes} provides statistics on the natural language instruction datasets.

\paragraph{\textsc{Lani} Dataset Collection Details}

The crowdsourcing process includes two stages.
First, a Mechanical Turk worker is shown a long, random trajectory in the overhead view and writes an instruction paragraph for a first-person agent.
The trajectories were generated with a sampling process biased towards moving around the landmarks.
Given the instruction paragraph, a second worker follows the instructions by controlling a discrete simple agent, simultaneously segmenting the instruction and trajectory into shorter segments. The output are pairs of instruction segments and discrete ground-truth trajectories. We restrict to a pool of workers who had previously qualified for our other instruction-writing tasks.

\paragraph{Demonstration Data}
We collect the demonstration datasets $\dataset^{\domainreal}$ and $\dataset^{\domainsim}$ by executing a hand-engineered oracle policy that has access to the ground truth human demonstrations, and collect observation data.
The oracle is described in Appendix~\ref{sec:app:oracle}.
$\dataset^{\domainsim}$ includes all instructions from original $\textsc{Lani}$ data, as well as the additional instructions we collect for our smaller environment. Due to the high cost of data collection on the physical quadcopter, $\dataset^{\domainreal}$ includes demonstrations on only 100 paragraphs of single-segment instructions, approximately 1.5\% of the data available in simulation.

\paragraph{Data Augmentation}
To improve generalization, we perform two types of data augmentation. First, we train on the combined dataset that includes both single-segment and two-segment instructions. Two-segment data consists of instructions and executions that combine two consecutive segments.
This increases the mean instruction length from 11.10 tokens to 20.97, and the mean trajectory length by a factor of 1.91.
Second, we randomly rotate the semantic map $\smworld$ and the gold visitation distributions by a random angle $\alpha \sim N(0, 0.5rad)$ to counter the bias of always flying forward, which is especially present in the single-segment data because of how humans segmented the original paragraphs.

\section{Additional Experimental Setup Details}
\label{sec:app:exp}

\subsection{Computing hardware and training time}
Training took approximately three days on an Intel i9 7980X CPU with three Nvidia 1080Ti GPUs. We used one GPU for the supervised learning process, one GPU for the RL process for both gradient updates and roll-outs, and one GPU for rendering simulator visuals. We ran five simulator processes in parallel, each at 7x real-time speed. We used a total of 400k RL simulation rollouts.

\subsection{Oracle implementation}
\label{sec:app:oracle}
The oracle uses the ground truth trajectory, and follows it with a control rule. It adjusts its angular velocity with a P-control rule to fly towards a dynamic goal on the ground truth trajectory. The dynamic goal is always 0.5m in front of the quadcopter's current position on the trajectory, until it overlaps the goal position. The forward speed is a constant maximum minus a multiple of angular velocity.

\subsection{Environment and Quadcopter Parameters}

\paragraph{Environment Scaling}
The original \textsc{Lani} dataset includes a unique 50x50m environment for each paragraph. Each environment includes 6--13 landmarks. Because the original data is for a larger environment, we scale it down to the same dimension as ours. We use the original data split, where environments are not shared between the splits.

\paragraph{Action Range}
We clip the forward velocity to the range $[0,0.7]$m/s and the yaw rate to $[-1.0,1.0]$rad/s. During training, we give a small negative reward for sampling an action outside the intervals $[-0.5,1.7]$m/s for forward velocity and $[-2.0,2.0]$rad/s for yaw-rate. This reduces the chance of action predictions diverging, and empirically ensures they stay mostly within the permitted range.

\paragraph{Quadcopter Safety}
We prevent the quadcopter from crashing into environment bounds through a safety interface that modifies the controller setpoint $\rho$. The safety mechanism performs a forward-simulation for one second and checks whether setting $\rho$ as the setpoint would cause a collision. If it would, the forward velocity is reduced, possibly to standstill, until it would no longer pose a threat. Angular velocity is left unmodified. This mechanism is only used when collecting demonstration trajectories and during test-time. No autonomous flight in the physical environment is done during learning.

\paragraph{First-person Camera}
We use a front-facing camera on the Intel Aero drone, tilted at a 15 degree pitch. The camera has a horizontal field-of-view of 84 degrees, which is less than the 90-degree horizontal FOV of the camera used in simulated experiments of \citet{blukis2018mapping}.

\section{Evaluation Details}
\label{sec:app:eval}

\subsection{Automated Evaluation Details}

We report two automated metrics: success rate and earth mover's distance (\textsc{EMD}).
The success rate is the frequency of executions in which the quadcopter stopped within 0.47m of the human demonstration stopping location.
To compute \textsc{EMD}, we convert the  trajectories executed by the quadcopter and the human demonstration trajectories to probability distributions with uniformly distributed mass across all positions on the trajectory.  \textsc{EMD} is then the earth mover's distance between these two distributions, using Euclidean distance as the distance metric.
\textsc{EMD} has a number of favorable properties, including: (a) taking into account the entire trajectory and not only the goal location, (b) giving partial credit for trajectories that are very close but do not overlap the human demonstrations, and (c) is smooth in that a slight change in the executed trajectory corresponds to at most a slight change in the metric.

\subsection{Human Evaluation Details}
\label{sec:app:human}

\paragraph{Navigation Instruction Quality}
One out of 73 navigation instructions that the majority of workers identified as unclear is excluded from the human evaluation analysis. The remaining instructions were judged by majority of workers not to contain mistakes, or issues with clarity or feasibility.

\begin{figure}[t]
\centering
\fbox{
\begin{minipage}{0.97\linewidth}
We need your help to understand how well our drone follows instructions. \\

Your task: Read the navigation instruction below, and watch the animation of the drone trying to follow the instruction. Then answer three questions. Consider the guidelines below:

\begin{itemize}
\item Looking around: The drone can only see what's directly in front of it as indicated by the highlighted region. Depending on the instruction, it may have to look for certain objects to understand where to go, and looking around for them is the right way to go and is efficient.
\item Bad instructions: If the instruction is unclear, impossible, or full of confusing mistakes, please indicate it by checking the checkbox. You must still answer all the questions - consider if the drone's behavior was reasonable given the bad instruction.
\item Objects: The drone observes the environment from a first-person view. The numbered images illustrate how each object would look like to the drone. Consider the appearance of objects from the first-person perspective in relation to the instructions.
\item Note: Try to answer each question in isolation, focusing on the specific behavior. For example, if the drone reached the goal correctly, but took the wrong path, you should answer "Yes, perfectly" for question 2, while giving a worse score for question 1. Similarly, if the drone went straight for the goal, that would be considered "efficient" behavior, even though it may have taken the wrong path to get there
\item The drone might sometimes decide not to do anything at all, maybe because it thinks it's already where it was instructed to go. If that happens you won't see any movement in the animation.
\item The drone always "stops" at the end, even if the motion appears to end abruptly.
\item The field is surrounded by a red fence on top, white fence on the right, blue fence on the bottom, and yellow fence on the left. The colorful borders are there to help you better distinguish between these colors.
\end{itemize}
\end{minipage}}
\caption{The instructions given to the crowdsourcing workers during human evaluation.}
\label{fig:eval-instructions}
\end{figure}

\paragraph{Mechanical Turk Evaluation Task}

\begin{figure*}[t]
\centering
\includegraphics[scale=0.29]{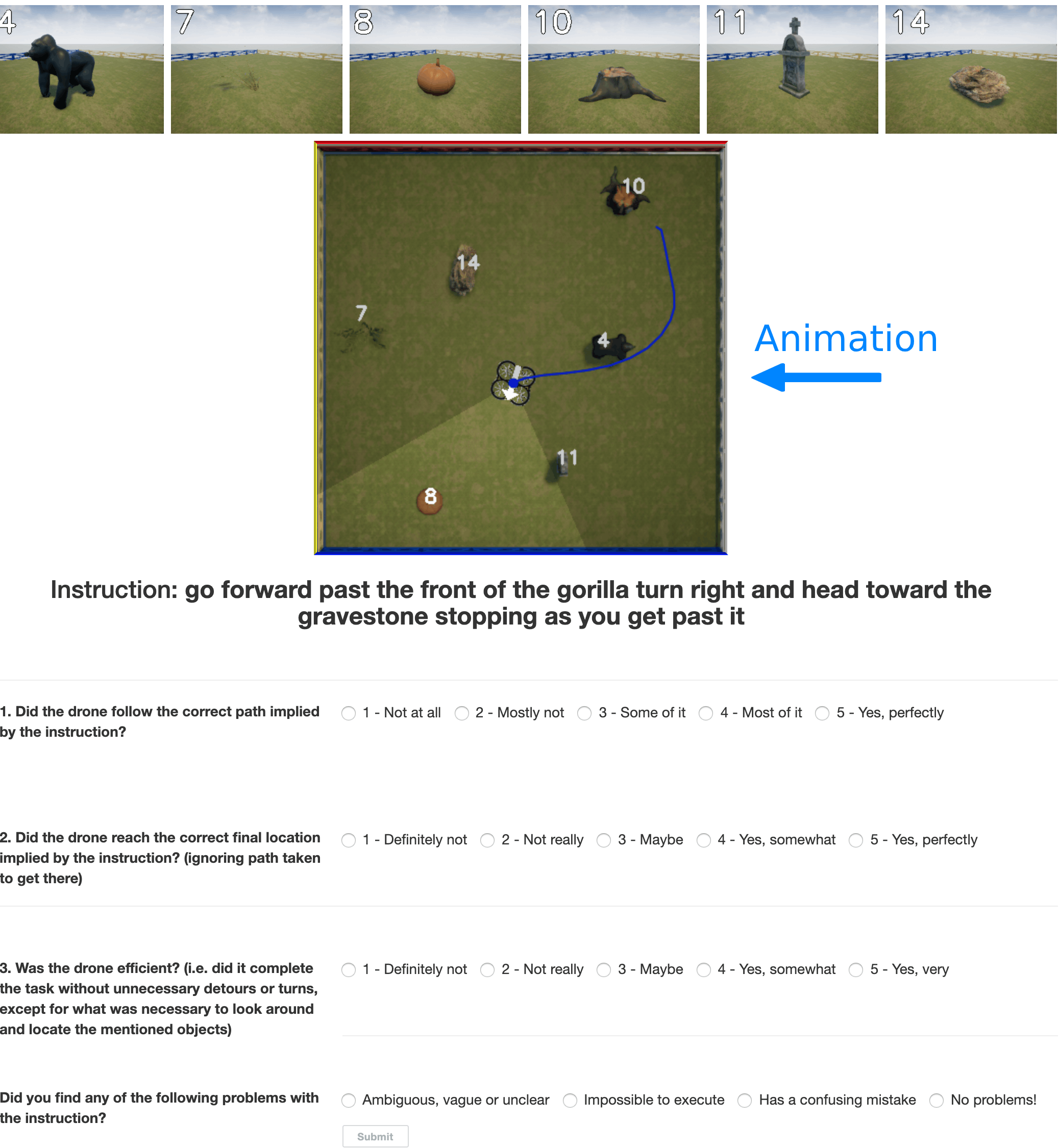}
\caption{Human evaluation task. The images on top show the various objects in the environment from a reasonable agent point of view, and numbers indicate correspondence to objects in the top-down view animation. The animation shows the trajectory that the agent took to complete the instruction. Because efficiency scores are unreliable, we do not report them.}
\label{fig:human:mturk}
\end{figure*}

\begin{figure*}[t]
\centering
\includegraphics[scale=0.35]{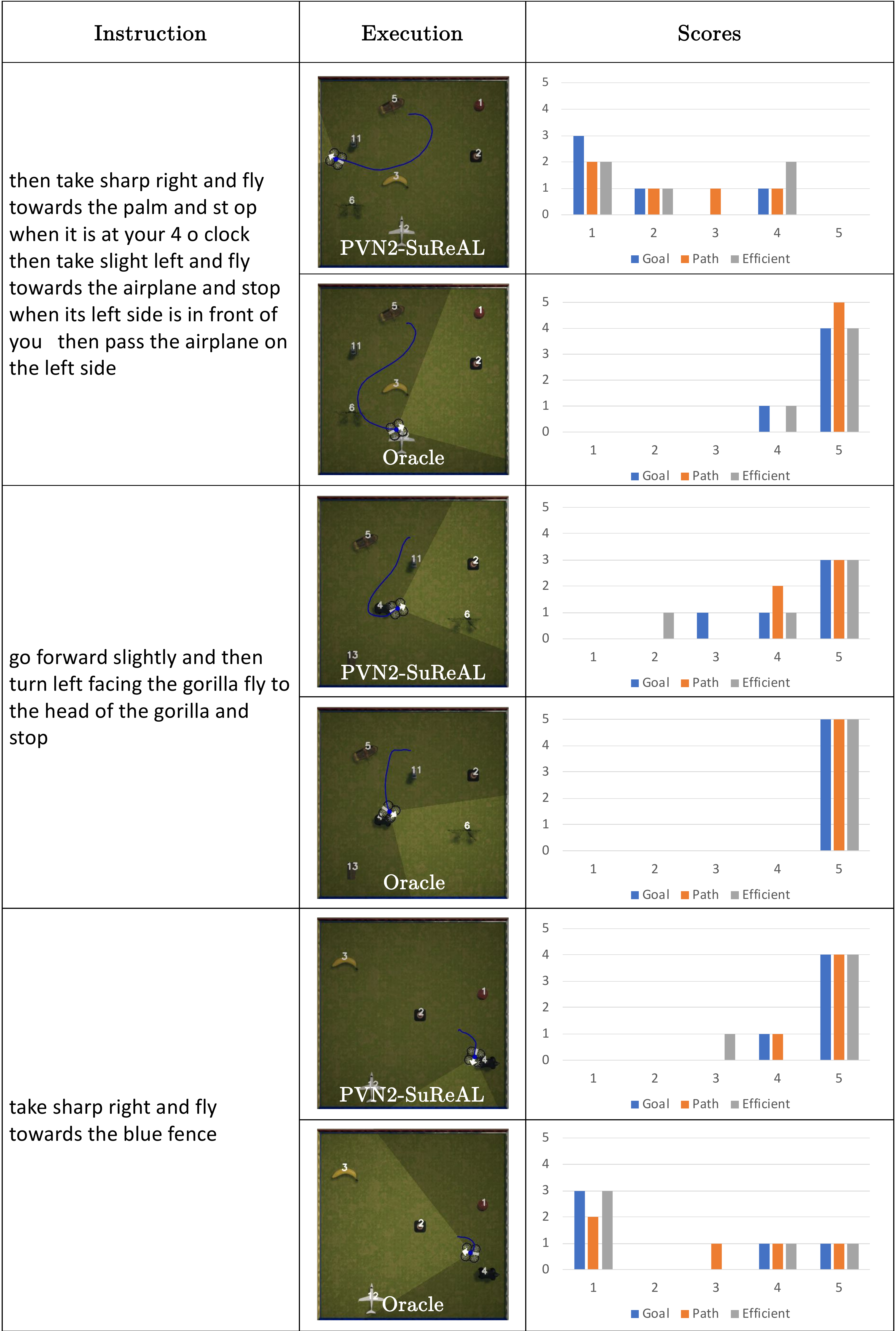}
\caption{Human evaluation scores for three instructions of different difficulty, for our model $\textsc{PVN2-\rail}$ and the $\textsc{Oracle}$. Horizontal axis represents Likert scores and bar heights represent score frequencies across five MTurk workers. Goal, Path, and Efficiency scores represent answers to the corresponding questions in Figure~\ref{fig:human:mturk}.}
\label{fig:human:scores}
\end{figure*}

Figure~\ref{fig:eval-instructions} shows the instructions given to workers for the human evaluation task.
Figure~\ref{fig:human:mturk} shows an example human evaluation task.
We use the simulated environment to visualize the quadcopter behavior to the workers because it is usually simpler to observe. However, we use this interface to evaluate performance on the physical environment, and use trajectories generated in the physical environment.
We avoid using language descriptions to describe objects to avoid biasing the workers, and allowing them to judge for themselves how well the instruction matches the objects. We observed that a large number of workers were not able to reliably judge the efficiency of agent behavior, since they generally considered correct behavior efficient and incorrect behavior inefficient. Therefore, we do not report efficiency scores.
Figure~\ref{fig:human:scores} shows examples of human judgements for different instruction executions.

\clearpage

\clearpage

\section{Additional Results}
\label{sec:app:results}

Table~\ref{tab:lani_results} shows simulation results on the full $\textsc{Lani}$ test and development data.

\begin{table}[t]
\footnotesize
\centering
\begin{tabular}{|c|c|c|c|c|c|c|c|}
\hline
\multirow{2}{*}{Method}   &\multicolumn{2}{|c|}{1-segment} & \multicolumn{2}{|c|}{2-segment}  \\
 \cline{2-8}
   & SR & EMD  & SR & EMD \\
\hline
\hline
\multicolumn{5}{|l|}{\textbf{Test Results On Full $\textsc{Lani}$ Test Set}} \\
\hline
\stopmodel              & \phantom{0}7.7   & 0.76   & \phantom{0}0.7   & 1.29 \\
\avgmodel              & \phantom{0}13.0   & 0.62   & \phantom{0}5.8   & 0.94 \\
\ddline{1-8}
\textsc{PVN-BC}        & \phantom{0}37.8   & 0.47   & \phantom{0}19.6   & 0.76 \\
\textsc{PVN2-BC}      & \textbf{\phantom{0}39.0}   & \textbf{0.43}   & \phantom{0}21.0   & 0.72 \\
\textsc{PVN2-$\rail$} & \phantom{0}37.2   & \textbf{0.43}   & \textbf{\phantom{0}21.5}   & \textbf{0.67} \\
\ddline{1-8}
\oraclemodel            & \phantom{0}98.6   & 0.15   & \phantom{0}93.9   & 0.20
 \\
\hline
\hline
\multicolumn{5}{|l|}{\textbf{Dev Results On Full $\textsc{Lani}$ Test Set}} \\
\hline
\textsc{PVN2-$\rail$}  & \phantom{0}35.8   & 0.44   & \phantom{0}19.8   & 0.68 \\
\textsc{PVN2-$\rail$-noExp} & \phantom{0}38.5   & 0.43   & \phantom{0}15.8   & 0.79 \\
\textsc{PVN2-$\rail$-noU} & \phantom{0}26.1   & 0.48   & \phantom{0}7.0   & 0.90 \\
\hline
\end{tabular}
\caption{Additional automated evaluation results on the full $\textsc{Lani}$ test and development sets of $50m\times50m$ environments, including the additional $4.7m \times 4.7m$ examples we added to support experiments on the real quadcopter. The 1-segment numbers in these tables are loosely comparable to prior work that used $\textsc{Lani}$~\cite{blukis2018mapping, misra2018mapping}, except in that we used additional data during training, and trained in a joint two-dataset domain-adversarial setting, which may have unpredictable effects on simulation performance. We also have reduced the camera horizontal FOV from 90 degrees to 84, which exacerbates observability challenges.}
\label{tab:lani_results}
\end{table}

\section{Common Questions}

\paragraph{Do you assume any alignment between simulated and real environments?}
Our learning approach does not assume that simulated and real data is aligned.

\paragraph{What is the benefit of $\rail$ over reinforcement learning with an auxiliary loss as a means of utilizing annotated data?}

There are a number of reasons to prefer $\rail$:
\begin{itemize}
    \item The 2-stage decomposition means that there is no gradient flow from second to first stage, and so the policy gradient loss will not update Stage 1 parameters anyway.
    \item With $\rail$, only the second stage needs to be computed during the PPO updates, which drastically improves training speed.
    \item In $\rail$, we do not need to send new Stage 1 parameters to actor processes at every iteration, since these parameters are not optimized with reinforcement learning.
\end{itemize}

\paragraph{Why did you select this particular set of baselines?}
The baselines have been developed for instruction following in a very similar environment to ours, allowing us to focus on evaluating domain transfer performance, exploration performance, and the effect of our learning method.
Instruction-following tasks have unique requirements for sample complexity and utilizing limited annotated data. Comparisons against general sim-to-real models may not yield useful insights, or yield insights that have already been demonstrated in prior work~\cite{blukis2018following}.

\paragraph{Why do you learn control, rather than using deterministic control?}
The predicted visitation distributions often have complex and unpredictable shapes, depending on the instruction and environment layout. Designing a hand-engineered controller that effectively follows the predicted distributions reduces to either (a) a challenging optimal control problem at test time or (b) a difficult engineering and testing challenge.
We instead opt to learn a solution at training time.
We find that learning Stage 2 behavior is more straightforward, and is in line with our general strategy of reducing engineering effort and use of hand-crafted representations.

\paragraph{How noisy are your pose estimates?}
We use a Vicon motion capture system to obtain pose estimates. The poses have sub-centimeter positional accuracy, but often arrive at the drone with delay due to network latency.

\paragraph{Is the model tolerant to noise?}
Prior work has tested the semantic mapping module that we use against noisy poses and found that it can tolerate moderate amounts of noise~\cite{blukis2018following} even without explicit probabilistic modelling. This is not the focus of this paper, but we find that our model is able to recover from erroneous pose-image pairs caused by network latency.
Evaluating the robustness of the model to more types of noise, including in pose estimates, is an important direction for future work.

\section{Hyperparameters}

Table~\ref{tab:hyper} shows the hyperparameter assignments. The hyperparameters were manually tuned on the development set to trade off use of computational resources and development time. We do not claim the selected hyperparameters are optimal, but we did observe that they consistently resulted in learning convergence and stable test-time behavior.

\begin{table*}[t]
  \footnotesize
  \centering
  \begin{tabular}{|c|c|}
  \hline
  Hyperparameter & Value \\
  \hline
  \hline
  \multicolumn{2}{|c|}{Environment Settings}\\
  \hline
Maximum yaw rate & $\velang_{\rm max} = 1m/s$\\
Maximum forward velocity & $\velfwd_{\rm max} = 0.7m/s$\\
  \hline
  \multicolumn{2}{|c|}{Image and Feature Dimensions}\\
  \hline
Camera horizontal FOV & $84^{\circ}$\\
Input image dimensions & $128 \times 72 \times3$\\
Feature map $\fmcam$ dimensions & $32 \times 18 \times 32$\\
Semantic map $\smworld$ dimensions & $64 \times 64 \times 32$\\
Visitation distributions $\stopvisit$ and $\trajvisit$ dimensions & $64 \times 64 \times 1$\\
Environment edge length in meters & $4.7m$\\
Environment edge length in pixels on $\smworld$ & $32$\\
  \hline
  \multicolumn{2}{|c|}{Model}\\
  \hline
Visitation prediction interval timesteps & $T_d = 1$\\
$\stopaction$ action threshold & $\kappa = 0.5$ \\
  \hline
  \multicolumn{2}{|c|}{General Learning}\\
  \hline
Deep Learning library & PyTorch 1.0.1\\
Classification auxiliary weight & $\lambda_{\rm percept} = 1.0$\\
Grounding auxiliary weight & $\lambda_{\rm ground} = 1.0$\\
Language auxiliary weight & $\lambda_{\rm lang} = 1.0$ \\
  \hline
  \multicolumn{2}{|c|}{Supervised Learning}\\
  \hline
Optimizer & ADAM \\
Learning Rate & $0.001$\\
Weight Decay & $10^{-6}$\\
Batch Size & $1$ \\
  \hline
  \multicolumn{2}{|c|}{Reinforcement Learning (PPO)}\\
  \hline
Num supervised epochs before starting RL ($K_{iter}^{B}$) & 25\\
Num epochs ($K_{\rm epoch}^{\rm RL}$) & 400\\
Iterations per epoch ($K_{\rm iter}^{\rm RL})$ & 50\\
Number of parallel actors & 4\\
Number of rollouts per iteration $N$ & 20\\
PPO clipping parameter & 0.1\\
PPO gradient updates per iter ($K_{\rm steps}^{\rm RL})$ & 8\\
Minibatch size & 2\\
Value loss weight & 1.0\\
Learning rate & 0.00025\\
Epsilon & 1e-5\\
Max gradient norm & 1.0\\
Use generalized advantage estimation & False\\
Discount factor ($\gamma$) & 0.99\\
Entropy coefficient & 0.001\\
Entropy schedule & multiply entropy coefficient by 0.1 after 200 epochs\\
  \hline
  \multicolumn{2}{|c|}{Reward Weights}\\
  \hline
  Stop reward weight ($\stopweight$) & 0.5\\
    Visitation reward weight($\visitweight$) & 0.3\\
    Exploration reward weight ($\exploreweight$) & 1.0\\
    Negative per-step reward ($\stepweight$) & -0.04\\
  \hline
  \end{tabular}
  \caption{Hyperparameter values.}
  \label{tab:hyper}
\end{table*}

\end{document}